\pdfoutput=1
\documentclass{article}
\usepackage[preprint]{neurips_2026}

\usepackage{amsmath,amsfonts,bm}

\def\eqref#1{equation~\ref{#1}}

\def\1{\bm{1}}

\DeclareMathAlphabet{\mathsfit}{\encodingdefault}{\sfdefault}{m}{sl}
\SetMathAlphabet{\mathsfit}{bold}{\encodingdefault}{\sfdefault}{bx}{n}

\usepackage{amsmath,amssymb,booktabs,graphicx,xcolor,url}
\usepackage{placeins}
\usepackage{hyperref}
\hypersetup{hidelinks}
\graphicspath{{figures/}}

\newcommand{\EE}{\mathbb E}
\newcommand{\PP}{\mathbb P}
\newcommand{\RR}{\mathbb R}
\newcommand{\CC}{\mathcal C}
\newcommand{\MM}{\mathcal M}
\newcommand{\norm}[1]{\left\lVert#1\right\rVert_2}
\newcommand{\ind}{\mathbf 1}
\DeclareMathOperator{\vMF}{vMF}
\title{The Geometry of Inference in Transformer Residual Streams}
\author{%
  Timur Mudarisov$^{1}$ \quad
  Mikhail Burtsev$^{2}$ \quad
  Radu State$^{1}$ \\[0.6em]
  \normalfont $^{1}$University of Luxembourg, Luxembourg \\
  \normalfont $^{2}$London Institute for Mathematical Sciences, London, UK
}

\begin{document}
\maketitle

\begin{abstract}
Transformer language models build predictions through successive residual updates, but how their representations become specific to an eventual outcome remains unclear. We study this process by comparing intermediate residual states with their own final states and an empirical bank of final states from other contexts. Across six pretrained language models, the own endpoint becomes preferable to the average alternative early, while many individual endpoints remain closer. These competing sets generally shrink with depth, but their membership changes and their surviving endpoints need not become more similar to one another. Directional alignment and endpoint rank can therefore improve while Euclidean distance to the final state changes little. We develop a simple high-dimensional model that separates the roles of norm, alignment, and endpoint geometry, showing how gradual directional changes can produce sharp reductions in competition. We also prove that a straight path toward the own endpoint cannot introduce new competitors under either Euclidean or cosine distance; observed entries thus establish departures from straight-line convergence. Finally, endpoints associated with lower-ranked output tokens tend to lie farther away in cosine distance across all studied models, connecting residual geometry to output organization. Together, these findings characterize increasing geometric specificity during transformer inference and explain why distance, competitor count, and concentration of the surviving endpoints provide distinct views of that process.
\end{abstract}

\section{Introduction}

Transformer language models build next-token predictions through successive residual updates. Prediction lenses show that information about eventual token predictions can often be decoded from intermediate states \citep{belrose2023tuned,ali2025entropy}, and studies of final representations identify output-related geometric structure \citep{park2025geometry}. These findings leave open how selectively an intermediate state distinguishes the final representation reached by its context from alternatives, and how these distinctions evolve across depth.

\begin{figure*}[t]
  \centering
  \includegraphics[width=0.95\textwidth]{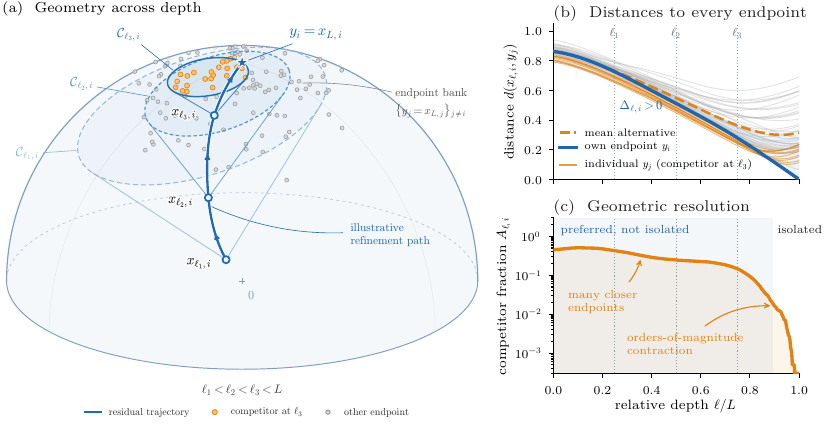}
  \caption{\textbf{Geometry of inference in residual streams.}
\textbf{(a)} Three intermediate states along a residual trajectory and an anisotropic bank of endpoints from other contexts. At layer $\ell$, alternatives closer to $x_{\ell,i}$ than the own endpoint $y_i$ form the competitor set $\mathcal{C}_{\ell,i}$. The competitor region is a ball under Euclidean distance and a spherical cap of directions under cosine distance.
\textbf{(b)} Distance to the own endpoint (blue), individual alternatives (gray, with competitors at $\ell_3$ highlighted in orange), and the mean alternative distance (dashed orange). The own endpoint is closer than the mean alternative from early depths ($\Delta_{\ell,i}>0$), while individual competitors remain.
\textbf{(c)} The competitor fraction $A_{\ell,i}$ remains significant after mean preference appears, then falls by orders of magnitude. Dotted lines in (b) and (c) mark the three states shown in (a). All panels use a synthetic three-dimensional model.}
  \label{fig:teaser}
\end{figure*}

We formulate the \emph{geometric inference hypothesis} that intermediate residual states express partial distinctions among possible final states and that residual updates refine those distinctions through changes in relative geometry (Figure~\ref{fig:teaser}). We operationalize this hypothesis using a fixed bank of \emph{residual endpoints}, the final measured residual states of sampled contexts. Each trajectory has an \emph{own endpoint}, and an alternative endpoint is a \emph{competitor} whenever it is closer to the current state than the own endpoint. The bank preserves contextual variation among final states, including states with the same top predicted token. Our contributions are as follows:

1. Across six pretrained models, we relate cosine endpoint distance to output-token rank and show that early average preference for the own endpoint coexists with many closer alternatives. Cosine competitor sets exhibit entries as well as exits over depths where their mean size declines, indicating that refinement can revise earlier endpoint comparisons. \\
2. We relate expected competitor counts to endpoint probability mass and show how gradual alignment can sharply reduce competition during a plateau in Euclidean distance. We also prove that monotone straight-line convergence yields nested competitor sets under Euclidean and cosine distance, so observed entries establish departures from this baseline.\\
3. We show that structured endpoint distributions fitted only to final states can predict mean competition along held-out trajectories more accurately than a uniform-sphere baseline.

\section{Geometry of Residual Inference in Language Models}
\label{sec:experiments}

\subsection{An Endpoint-Based Probe of Inference}
\label{sec:setup}

To measure refinement relative to alternatives, we need a reference population against which each intermediate state can be compared. Final residual states provide such a population in the model's native residual coordinates \citep{elhage2021framework}. The own endpoint records where a computation ends in a particular token position, while endpoints from other contexts describe other final representations reached by the same model. The analysis considers the observed trajectory and asks when and how it becomes specific to its own endpoint. %

For context $i$, let $x_{\ell,i}\in\RR^d$ be the residual state at the final input position after block $\ell$, before final normalization, and let $y_i=x_{L,i}$ be its own endpoint. We study Gemma-2B and Gemma-7B \citep{gemmateam2024gemma}, Qwen2.5-1.5B and Qwen2.5-7B \citep{qwenteam2024qwen25}, Mistral-7B \citep{mistralai2024mistralv03}, and Llama-3-8B \citep{aimeta2024llama3card} on 1,024 non-overlapping 256-token contexts per model from FineWeb \texttt{sample-10BT} \citep{penedo2024fineweb}. The primary bank contains all other endpoints, so each query has $M_i=1{,}023$ alternatives. Appendix~\ref{app:protocol} gives the details for endpoint modeling.

\textbf{Measuring specificity relative to alternatives.}
Distance to the own endpoint alone cannot tell whether an intermediate state distinguishes it from other endpoints that are also nearby. We therefore compare each alternative with the own endpoint under Euclidean and cosine distance,
\begin{equation}
 d_E(x,y)=\norm{x-y},\qquad
 d_C(x,y)=1-\frac{x^\top y}{\norm{x}\norm{y}}.
 \label{eq:metrics}
\end{equation}
Euclidean trajectory distances are divided by $R_i=\norm{y_i}$, preserving their ordering for a fixed state. Cosine comparisons use nonzero states and endpoints. For an eligible bank $\mathcal B_i$ of size $M_i$, the competitor set and competitor fraction are
\begin{equation}
 \CC^{(m)}_{\ell,i}=\{j\in\mathcal B_i:d_m(x_{\ell,i},y_j)<d_m(x_{\ell,i},y_i)\},
 \qquad A^{(m)}_{\ell,i}=\frac{|\CC^{(m)}_{\ell,i}|}{M_i}.
 \label{eq:cset}
\end{equation}
The own endpoint's strict retrieval rank is $1+|\CC^{(m)}_{\ell,i}|$, with ties excluded. A smaller competitor fraction thus indicates greater \emph{geometric specificity} to the own endpoint within the bank.

\textbf{Why endpoint geometry is relevant to prediction.}
\label{sec:exp_modelrelevance}
To connect endpoint geometry to prediction, we ask whether more probable next tokens correspond to closer endpoints. For each context, we rank tokens by their final output probabilities. For each ranked token, we identify other contexts in which that token is the model’s most probable next token and measure the mean distance from their endpoints to the original context’s endpoint.

\begin{figure}[t]
 \centering
 \includegraphics[width=\linewidth]{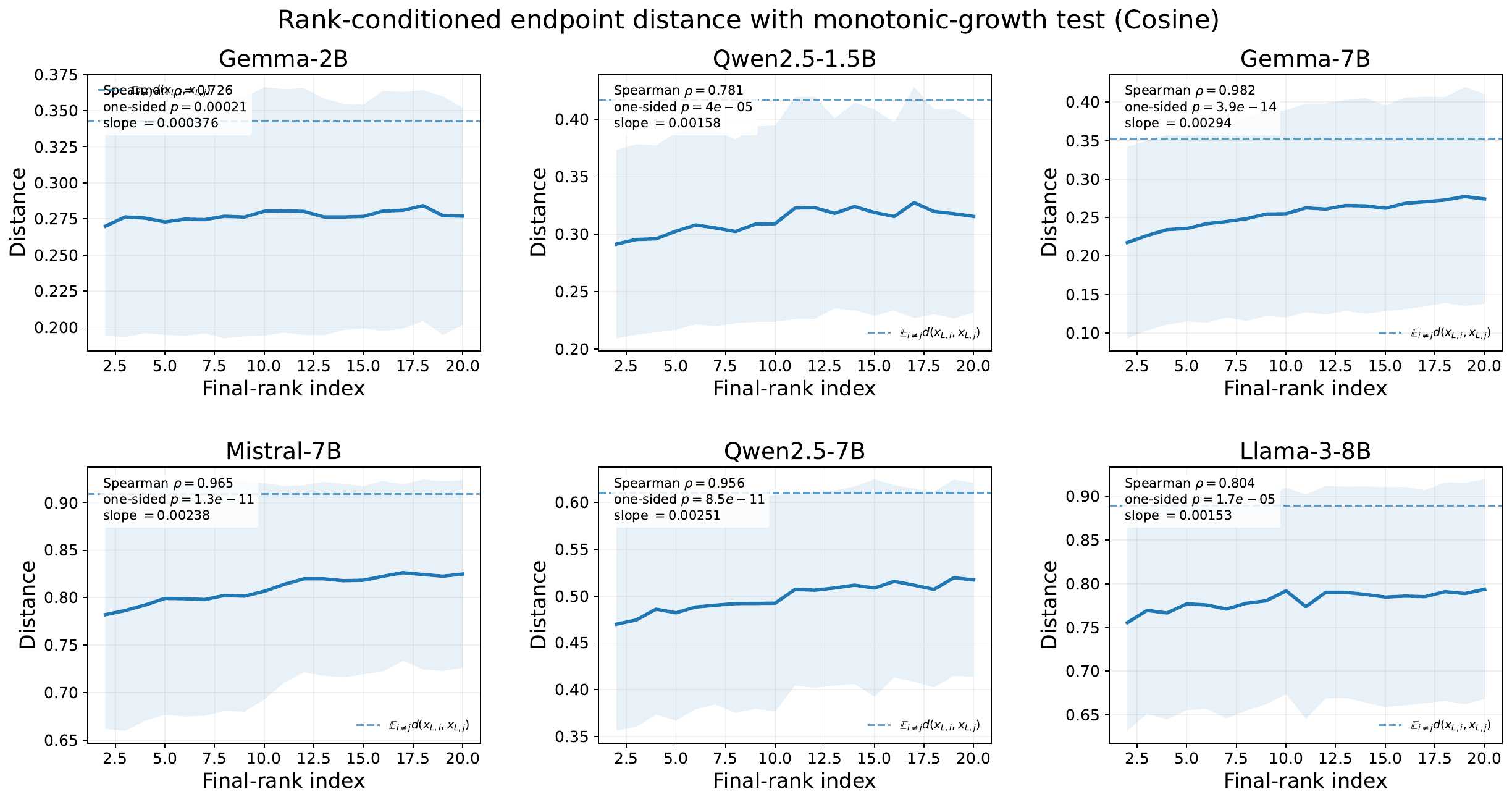}
\caption{\textbf{Less probable tokens correspond to more distant endpoints.} Mean cosine distance from the query's endpoint to endpoints from other contexts where the query's rank-$k$ token is the most probable next token. Distance tends to increase with $k$, with positive trends in the reported one-sided tests for all six models. Horizontal lines show unrelated-endpoint baselines.}
 \label{fig:rank}
\end{figure}

Formally, the output distribution is $p_i=\operatorname{softmax}(W_U N_f(y_i))$, where $N_f$ is the final normalization. Let $v_{i,k}$ be its rank-$k$ token and $J_{i,k}$ the alternative endpoints whose top prediction is $v_{i,k}$. For nonempty groups, we measure
\begin{equation}
 \delta_i^{(m)}(k)=\frac{1}{|J_{i,k}|}\sum_{j\in J_{i,k}}d_m(y_i,y_j).
 \label{eq:rankdist}
\end{equation}
Across all six models, endpoints whose top predictions receive lower probability under the query context tend to lie farther from the query’s own endpoint in cosine distance (Figure~\ref{fig:rank}). Endpoints sharing a top prediction are also more compact on average, and greater endpoint distance is associated with greater divergence between output distributions (Appendix~\ref{app:modelrelevance}). These associations support using endpoint geometry to study inference, although geometry explains only part of the variation in model outputs. We interpret the competitor fraction as a measure of geometric specificity within the endpoint bank, without treating it as a calibrated token probability.

\subsection{Early Preference with Many Competitors}
\label{sec:exp_preference}\label{sec:exp_contraction}

\begin{figure}[t]
 \centering
 \includegraphics[width=0.96\linewidth]{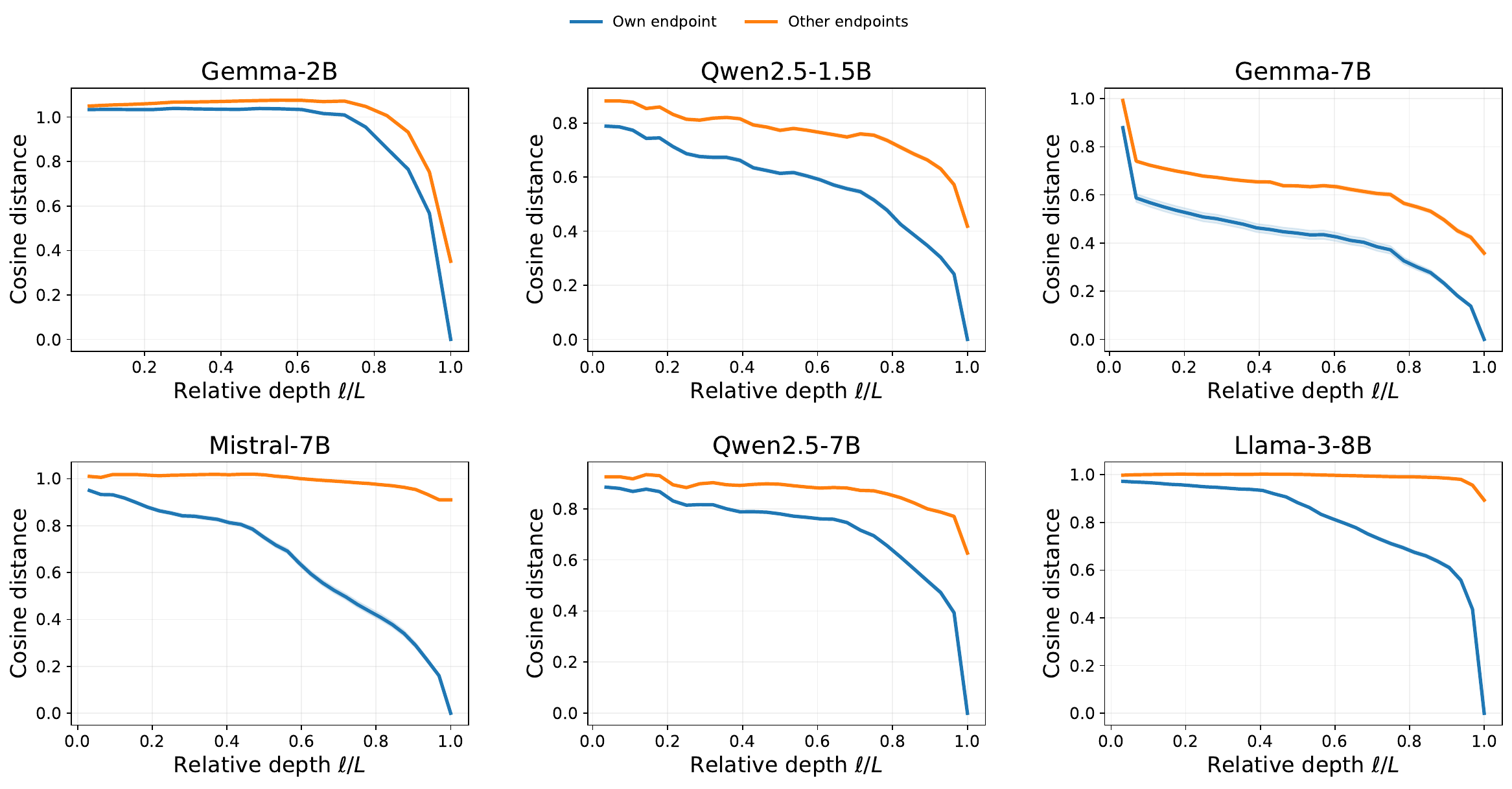}\par
 {\small\textbf{(a) Distance to the own endpoint and mean alternative}}\par\vspace{0.5em}
 \includegraphics[width=0.96\linewidth]{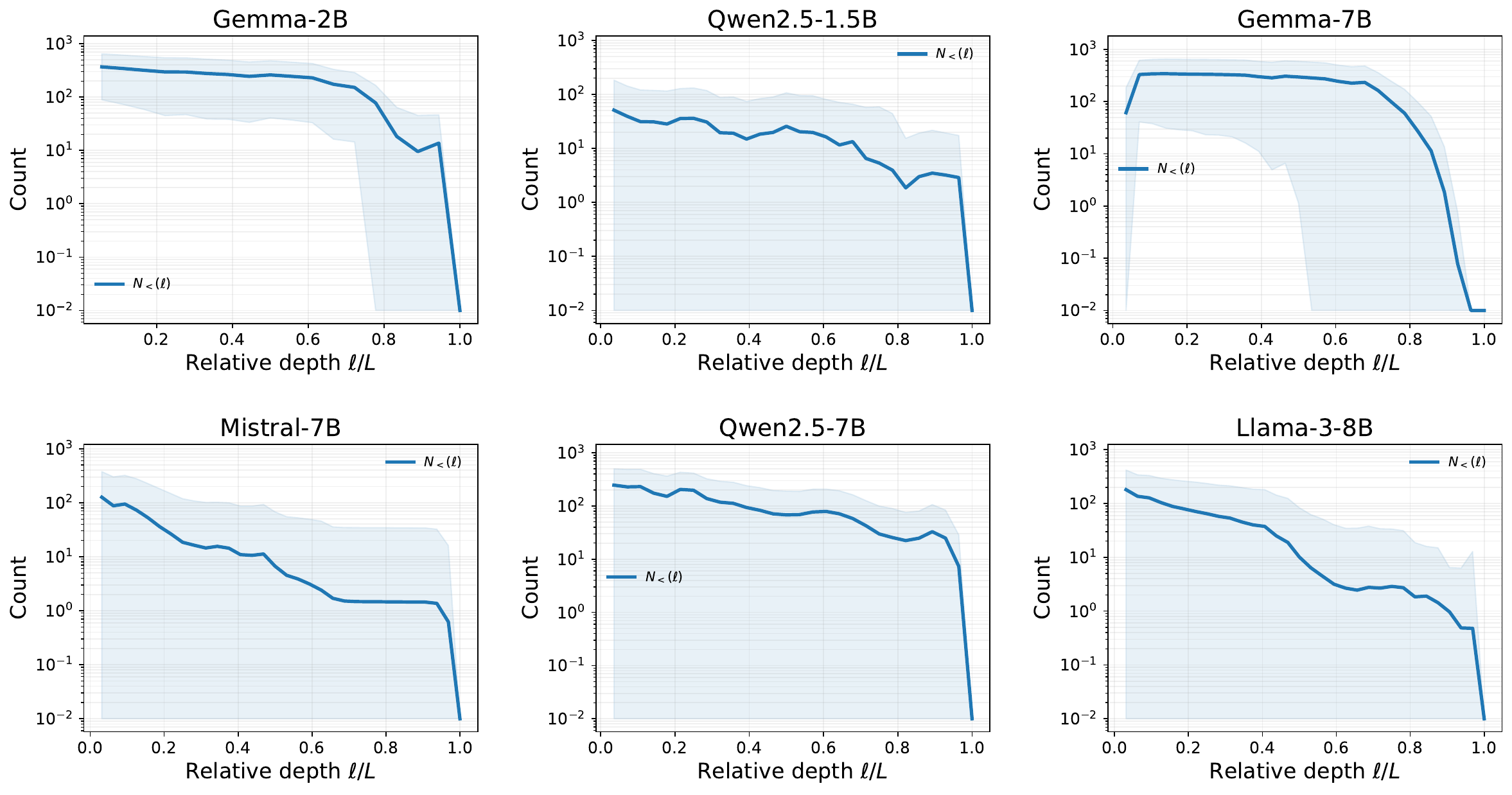}\par
 {\small\textbf{(b) Number of strict competitors}}
 \caption{\textbf{Early average preference leaves many individual competitors.} The own endpoint is closer than the average alternative before the competitor count becomes small. The paired measurements distinguish the presence of an endpoint preference from its selectivity within the bank. Cosine counts generally decrease with depth, with architecture-dependent fluctuations. Count shading gives across-context spread, and zero is displayed at $10^{-2}$ on the logarithmic axis. %
 }
 \label{fig:trajectory}\label{fig:contraction}
\end{figure}

With an endpoint bank estimating output geometry of the model, we can ask how early the residual state begins to distinguish its own endpoint and how selective that distinction is. We use two complementary comparisons. Mean distance to alternative endpoints detects an overall preference for the own endpoint, while competitor count records how many individual alternatives remain closer. The average-preference margin is
\begin{equation}
 \Delta^{(m)}_{\ell,i}=\frac{1}{M_i}\sum_{j\in\mathcal B_i}d_m(x_{\ell,i},y_j)-d_m(x_{\ell,i},y_i).
 \label{eq:delta}
\end{equation}
Positive $\Delta^{(m)}_{\ell,i}$ means the own endpoint is closer than the average alternative. Across all six models and both metrics, the document-averaged margin is positive from the earliest measured post-block state (Appendix~\ref{app:separation}). Many individual competitors nevertheless remain after this preference is detectable (Figure~\ref{fig:trajectory}). Under Euclidean distance, five models retain hundreds of competitors across substantial portions of their depth (Appendix~\ref{app:metriccompanions}).

This joint pattern is central to the inference hypothesis. Intermediate representations exhibit a measurable preference for their own endpoints while many individual alternatives remain closer. The subsequent reduction in competitors describes increasing selectivity of that preference. Its timing and strength vary across architectures.

\subsection{Competitor Sets Shrink with Changing Membership}
\label{sec:turnover}

The two metrics also show why distance to the own endpoint gives an incomplete account of refinement. Cosine competition often decreases earlier than Euclidean competition, and directional alignment can improve while Euclidean endpoint distance remains nearly constant, particularly in intermediate layers of Mistral-7B and Llama-3-8B. A Euclidean plateau can therefore accompany progress in directional specificity. The cosine profiles also persist within sampling variation when the endpoint bank is uniformly subsampled (Appendix~\ref{app:banksize}).

\textbf{Input token and context contribute to the early geometric preference.}
We test whether early preference for the own endpoint can be explained by shared document content or the final input token. Excluding endpoints from the query's document leaves the cosine preference and competitor-fraction curves essentially unchanged. Restricting alternatives to contexts ending in the same input token preserves preference from the first measured layer, although margins are smaller and more competitors remain. This pattern indicates that token identity contributes to early preference. Removing each state's component along its input-token embedding direction also preserves first-layer cosine preference in all six models. Together, these controls support a contribution from the broader context, while leaving information carried through residual connections and token identity encoded in other directions as possible contributors (Appendix~\ref{app:tokencontrols}).

\begin{figure}[t]
 \centering
 \includegraphics[width=\linewidth]{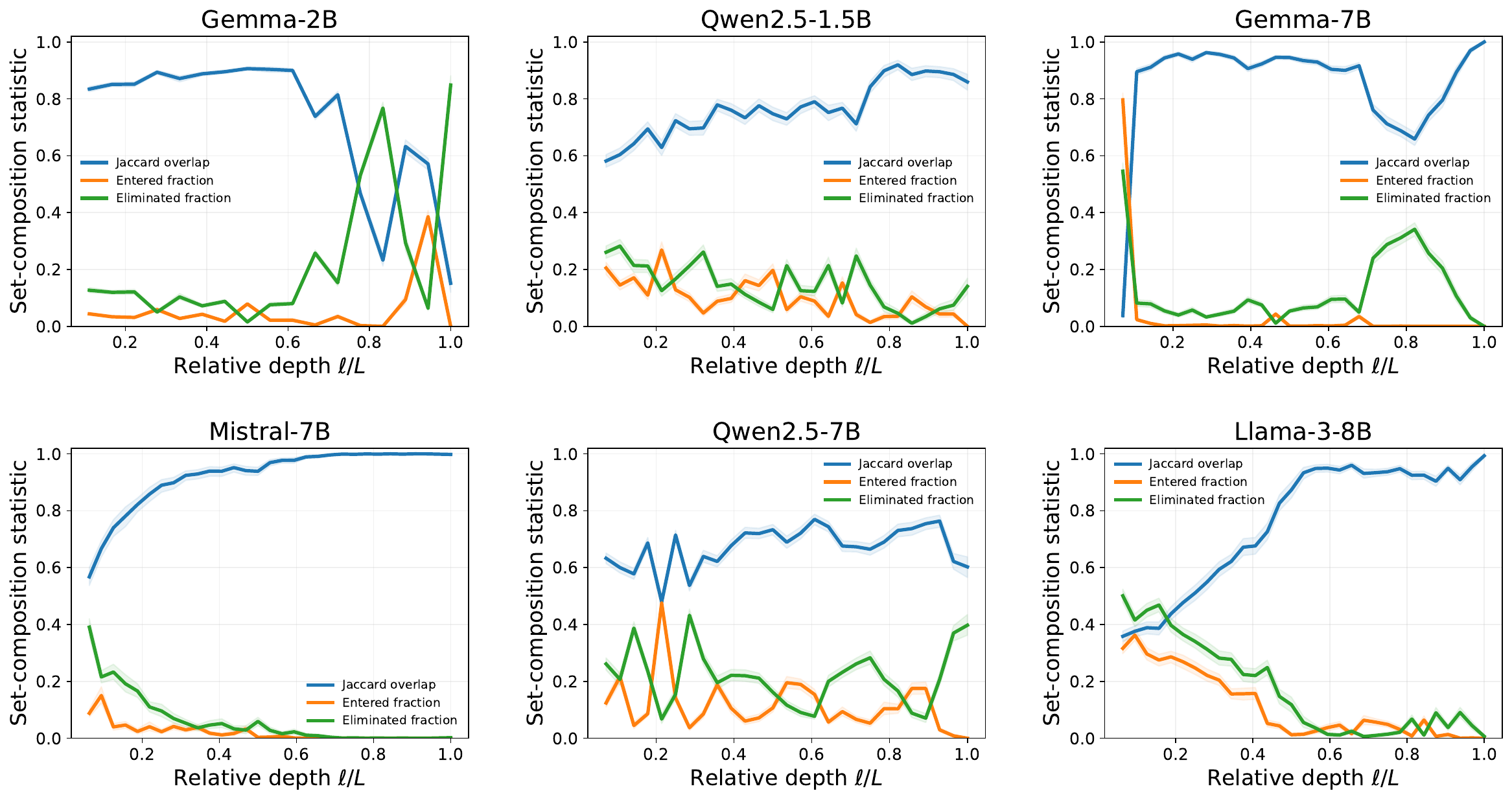}
 \caption{\textbf{Falling counts can conceal changes in competitor identity.} Adjacent-layer Jaccard overlap and fractions of entering and exiting endpoints under cosine distance. Entries show that refinement can change which alternatives compete, excluding a universally nested elimination process. Late overlap values can include empty-to-empty transitions, assigned Jaccard value one.}
 \label{fig:composition}
\end{figure}

A decreasing count of competitors from layer to layer (Fig.~\ref{fig:contraction}) can arise through successive removal from a fixed collection of competitors or through a process in which some alternatives enter as others leave. These possibilities give different accounts of refinement. Successive removal would preserve every earlier exclusion, whereas changing membership would allow the state to alter which endpoints it favors along the trajectory. To track composition of competitor sets across adjacent layers, we measure Jaccard overlap together with entries $\CC_{\ell,i}\setminus\CC_{\ell-1,i}$ and exits $\CC_{\ell-1,i}\setminus\CC_{\ell,i}$. Entry and exit fractions are normalized by the current and previous set sizes, respectively, before averaging across contexts. Figure~\ref{fig:composition} shows both entries and exits under cosine distance, including sustained turnover in Qwen2.5-1.5B. Euclidean sets are closer to nested over much of the trajectory and change mainly near the end (Appendix~\ref{app:turnover}).

The cosine result shows that increasing specificity can involve a changing collection of geometric alternatives. Endpoints that did not compete at one layer can compete at the next, and these entries occur over depths where the mean competitor count decreases. Entries include both first-time entries and possible returns, which the statistic does not distinguish. 

There is a further distinction between fewer competitors and a tighter group of competitors. Their count depends on comparisons with the residual state, whereas their cohesion depends on distances to one another. A shrinking set can become less cohesive if the removed endpoints were especially close to the remaining ones. Appendix~\ref{app:cohesion} gives an exact fixed-bank construction in which the count falls while mean pairwise Euclidean and cosine distances both increase. This mathematical example separates set size from concentration. We use shrinkage to describe the number of competitors, without inferring an empirical cohesion trend from the turnover or whole-layer spread measurements.

\section{Simple Models of Residual Trajectories}
\label{sec:theory}

The empirical results require an account of three different aspects of refinement: changes in competitor membership, improvements in directional specificity during distance plateaus, and large reductions in competitor count. 

\textbf{How an update changes endpoint preference.}
To explain entries and exits, it is useful to express each comparison as a signed margin. A negative margin identifies a competitor, and crossing zero changes membership. Writing $u_j=y_j/\norm{y_j}$, define the squared Euclidean margin and the unnormalized directional margin as
\begin{equation}
 m^E_{ij}(x)=\norm{x-y_j}^2-\norm{x-y_i}^2,
 \qquad m^C_{ij}(x)=x^\top(u_i-u_j).
 \label{eq:update_margins}
\end{equation}
For a residual update $\delta$, these margins change by
\begin{equation}
 m^E_{ij}(x+\delta)-m^E_{ij}(x)=2\delta^\top(y_i-y_j),
 \qquad
 m^C_{ij}(x+\delta)-m^C_{ij}(x)=\delta^\top(u_i-u_j).
 \label{eq:margin_changes}
\end{equation}
An update increases the corresponding margin when it projects positively onto the endpoint-difference direction. The cosine-distance gap equals $m^C_{ij}(x)/\norm{x}$, so its sign agrees with the directional margin although its magnitude also depends on norm. Because difference directions vary across the bank, one update can increase some margins and decrease others. Thus residual movement changes several endpoint preferences together, potentially removing some competitors while introducing others.

\textbf{A straight-path baseline for turnover.}
The simplest convergence path places a stronger constraint on these comparisons. Let $x(t)=(1-t)x_0+t y_i$, with $t$ increasing from zero to one and the endpoint bank fixed. Under either metric, its competitor sets are nested. Both margins in Eq.~\ref{eq:update_margins} are affine in $t$ and nonnegative at $t=1$, so any alternative with a nonnegative margin remains outside the competitor set thereafter. Cosine comparisons require nonzero states and endpoints. Appendix~\ref{app:theory} gives the full proof. Thus the endpoint entries in competitor set (Fig.~\ref{fig:composition}) establish departures from monotone straight-line convergence on the affected trajectories.

\textbf{Why directional progress can coexist with flat distance.}
The observed distance plateaus raise a different issue: Euclidean distance combines changes in direction with changes in scale. Let $r=\norm{x}$, $R=\norm{y_i}$, and $c=(x/r)^\top(y_i/R)$. Then
\begin{equation}
 \frac{\norm{x-y_i}^2}{R^2}
 =\left(\frac rR\right)^2+1-2\frac rR c.
 \label{eq:norm_alignment_distance}
\end{equation}
Increasing alignment can be offset by changing norm. Along a path with increasing $c$ and $r/R=2c$, normalized distance remains one for $0<c<1/2$ even as the direction becomes better aligned with the own endpoint. This shows that a flat Euclidean curve is compatible with the directional refinement.

\textbf{Endpoint distribution and competitor count.}
To explain changes in competitor count, we must consider both the movement of residual state and the distribution of alternative endpoints. We quantify their combined effect through the probability that a sampled endpoint is closer to the state than the own endpoint. Let $Y\sim P_Y$ be an alternative endpoint and $U=Y/\norm{Y}$ its direction. For a fixed state $x_{\ell,i}$ and own endpoint $y_i$, define
\begin{align}
q^E_{\ell,i}=\PP!\left(\norm{Y-x_{\ell,i}}<\norm{y_i-x_{\ell,i}}\right),\quad
q^C_{\ell,i}=\PP!\left(u_{\ell,i}^\top U>u_{\ell,i}^\top u_i\right),
\label{eq:capmass}
\end{align}
where $u_{\ell,i}=x_{\ell,i}/\norm{x_{\ell,i}}$ and $u_i=y_i/\norm{y_i}$. Under Euclidean distance, competitors lie inside the ball centered at the residual state whose boundary passes through the own endpoint. Under cosine distance, their directions lie in a spherical cap containing directions better aligned with the state than the own endpoint. We call the probability assigned to either region its \emph{competitive mass}. For $M$ alternatives with distribution $P_Y$, the expected competitor count is $Mq^{(m)}_{\ell,i}$. The trajectory determines how these regions change, while the endpoint distribution determines how much competitive mass is gained or lost.

\textbf{From gradual alignment to sharp count reductions.}
A simple high-dimensional model illustrates why average preference can coexist with many competitors and how gradual alignment can sharply reduce their number. We decompose endpoints and intermediate states into a shared direction and an orthogonal subspace:
\begin{equation}
y_j=R\left(\rho m+\sqrt{1-\rho^2}\,v_j\right),
\qquad
x=r\left(bm+\sqrt{1-b^2}\,w\right).
\label{eq:shared_axis_model}
\end{equation}
Here $R,r>0$ are the endpoint and intermediate-state norms, $m$ is a shared unit direction, and $0\leq\rho<1$ and $|b|<1$ control alignment with $m$. The unit vectors $v_j$ and $w$ lie in the $D$-dimensional subspace orthogonal to $m$, with $D\geq2$. Alternative directions $v_j$ are uniform on its sphere. Alignment with the own endpoint in this subspace is $a=w^\top v_i$. Because all endpoints have the same norm and shared component, an alternative competes exactly when $w^\top v_j>a$.

For $M$ independent alternatives, the expected competitor count is
\begin{equation}
\EE[N_<\mid x,y_i]=M\overline F_D(a),
\qquad \overline F_D(a)=\PP(w^\top V>a),
\quad V\sim\mathrm{Unif}(\mathbb S^{D-1}).
\label{eq:spherical_tail_count}
\end{equation}
Any positive $a$ makes the own endpoint closer in cosine distance than a random alternative on average. However, when $0<a\ll D^{-1/2}$, the expected competitor fraction remains close to one half. An expected count much smaller than one requires $\overline F_D(a)\ll1/M$. In high dimension, $\overline F_D(a)\approx1-\Phi(\sqrt D,a)$ near $a=0$, where $\Phi$ is the standard normal CDF. This approximation shows how small changes in alignment can substantially reduce competitive mass.

The rate of count reduction depends on both alignment speed and the distribution of alternatives near the comparison boundary:
\begin{equation}
-\frac{d}{dt}\log\EE[N_<(t)]
=\frac{f_D(a(t))}{\overline F_D(a(t))}\,a'(t),
\label{eq:contraction_rate}
\end{equation}
where $f_D=-\overline F_D'$ is the density of the alternative score $w^\top V$ (Appendix~\ref{app:theory}). The ratio $f_D(a)/\overline F_D(a)$ measures the relative sensitivity of competitor count to alignment, allowing smooth directional changes to produce sharp count reductions.

These reductions can also occur while Euclidean distance remains constant. Setting $b=\rho=0$ and $r/R=2a$ keeps $\norm{x-y_i}/R=1$ as $a$ increases within $(0,1/2)$. The construction therefore shows how average preference, endpoint distance, and competitor count can evolve differently. The depth at which counts fall depends on the prescribed alignment trajectory. This motivates testing the contribution of endpoint geometry along fixed, observed residual trajectories.

The preceding model relates competitor count to the distribution of alternative endpoints. We now test this relationship using real endpoint populations, whose directional structure varies across architectures (Appendix~\ref{app:ambient}). We ask whether fitted endpoint distributions predict mean competitor fractions along held-out trajectories when the intermediate states and own endpoints remain fixed.

\textbf{Fitting endpoint distributions.}
We compare five families with different assumptions about directional structure. The uniform sphere provides an isotropic baseline, while the spherical cap and single von Mises--Fisher (vMF) distribution concentrate endpoints around one preferred direction. A vMF mixture represents multiple directional modes, and a projected-normal model captures anisotropic variation by normalizing samples from a Gaussian fitted to the mean and covariance of unit endpoints \citep{mardia2000directional,banerjee2005vmf,wang2013projected}. All families use the same empirical distribution of endpoint norms, sampled independently of direction, so differences in their predictions reflect the fitted directional structure.

Fitting and model selection use only final endpoints. We split documents approximately 60/20/20 into training, validation, and test sets, select complexity within each family using four endpoint-distribution diagnostics, and refit on the combined training and validation endpoints. The best held-out diagnostic scores improve over the sphere by factors of $4.6$--$7.5$ (Appendix~\ref{app:models}). To determine whether these gains extend to the regions relevant to inference, we next test how accurately the fitted distributions predict competitive mass along held-out residual trajectories.

\textbf{Predicting competitor fractions.}
We sample an alternative bank of $B=256$ endpoints from each fitted distribution and evaluate it along held-out residual trajectories, keeping the intermediate states and own endpoints fixed. The predicted mean competitor fraction at layer $\ell$ is
\begin{equation}
 \widehat A^{(m)}_{\MM}(\ell)=\frac{1}{N_{\rm test}B}\sum_{i,b}
 \ind\!\left[d_m(x_{\ell,i},\widetilde Y_b)<d_m(x_{\ell,i},y_i)\right].
 \label{eq:predA}
\end{equation}
where $N_{\rm test}$ is the number of held-out contexts and $\widetilde Y_b$ are the sampled endpoints. We keep each bank fixed across depth and average the resulting curves over eight banks. The comparison evaluates how accurately the fitted distribution captures competitive mass along previously unseen trajectories.

\begin{figure}[t]
\centering
\includegraphics[width=0.95\linewidth]{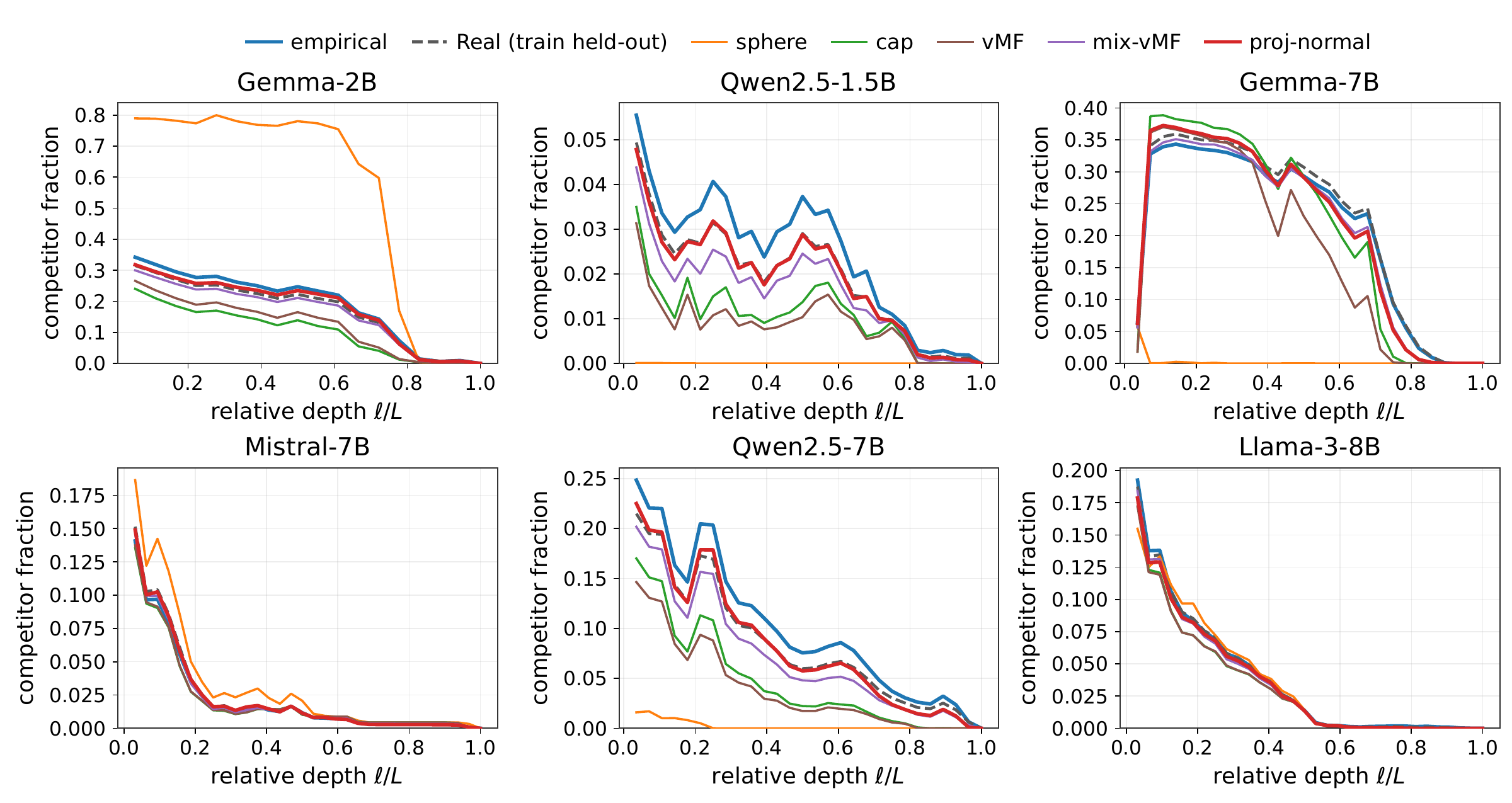}
\caption{\textbf{Predicting competition from endpoint geometry.} Mean cosine competitor fractions along held-out trajectories, comparing the empirical test bank with banks sampled from fitted distributions or training endpoints. Intermediate states and own endpoints are fixed across comparisons. Sampled banks remain fixed across depth, with estimates averaged over eight banks.}
\label{fig:predicted}
\end{figure}

The empirical competitor fractions use the test endpoint bank, excluding each context's own endpoint. Banks of $B$ training endpoints provide a reference based on sampled real endpoints. Appendix~\ref{sec:methods} details the differences in sampling pools and document overlap between these banks.

Competitor fractions span several orders of magnitude, so we compare predicted and empirical curves using the mean absolute difference on a base-10 logarithmic scale:
\begin{equation}
 \mathrm{err}_{\log}=\frac{1}{L-1}\sum_{\ell<L}\left|\log_{10}\!\big(\widehat A_{\MM}(\ell)+\varepsilon\big)-\log_{10}\!\big(A(\ell)+\varepsilon\big)\right|,\qquad \varepsilon=10^{-3},
 \label{eq:logerr}
\end{equation}
Here $A(\ell)$ is the mean empirical competitor fraction, and the offset $\varepsilon$ keeps the logarithm defined at zero. We exclude the final layer, where competitor fractions are zero by construction.

The projected-normal model gives the most consistently accurate predictions of cosine competitor fractions, achieving the lowest synthetic error in five of six models (Figure~\ref{fig:predicted}, App.~\ref{app:models} Table~\ref{tab:traj_log}). The vMF mixture is slightly better for Gemma-7B and ranks second elsewhere. Both outperform the cap and single vMF models across all six architectures, while the uniform sphere has the largest error in five. 
These results show that capturing directional structure improves predictions of how many alternatives remain competitive along observed residual trajectories.

\section{Discussion and Conclusion}
\label{sec:discussion}

Prior work shows that intermediate residual states contain information about eventual token predictions \citep{geva2022transformer,belrose2023tuned} and that final representations have output-related geometric structure \citep{park2025geometry}. To study how a trajectory becomes specific to its own endpoint, we use final states from other contexts as a reference population. Endpoints whose top predictions are less probable under the query context tend to lie farther from its endpoint, linking endpoint distance to prediction. Consistent with earlier evidence of predictive information in intermediate states, the document-averaged margin favors the own endpoint from the earliest layers in all six models, although many individual alternatives remain closer. The later reduction in competitor counts shows how this early average preference develops into greater selectivity among contextual endpoints.

Entropy-Lens identifies expansion and pruning in vocabulary-level predictions \citep{ali2025entropy}, and prior analyses track changes in representation neighborhoods across depth \citep{valeriani2023geometry,jiang2026layerwise}. Our fixed endpoint bank lets us ask whether an alternative that was farther than the own endpoint becomes closer at a later layer. Cosine competitor sets show endpoint entries as well as exits over depths where their mean count decreases. Euclidean sets are closer to nested over much of the trajectory and change mainly near the end. The margin equations explain how one residual update can strengthen preference over some endpoints while weakening it over others. A straight path toward the own endpoint would produce nested sets under either metric, so observed entries establish departures from monotone straight-line convergence on the affected trajectories. Increasing endpoint specificity can therefore involve revisions to earlier comparisons, with some alternatives becoming competitive as others cease to compete.

Competitor counts also depend on the distribution of alternative endpoints. Earlier work documents anisotropy and prediction-related structure in residual representations \citep{ethayarajh2019contextual,xu2026scale,guda2026stratification,lombardo2026geometric}, while semantic reference frames formalize trajectories using vocabulary-derived anchors \citep{gu2026semrf}. In our framework, the current state and own endpoint define a comparison ball or directional cap whose probability mass determines the expected competitor count. The shared-axis model shows how a small positive alignment can favor the own endpoint over the average alternative while leaving many competitors, and how gradual alignment can later produce a sharp reduction in their number. Norm changes also permit directional progress during a plateau in Euclidean distance. To test the role of population geometry, we fit endpoint distributions independently of the trajectories and evaluate their predictions along held-out paths, keeping intermediate states and own endpoints fixed. The projected-normal family predicts the cosine competitor curves most accurately in five of the six models, and both it and the vMF mixture outperform the uniform-sphere baseline. Endpoint geometry thus provides a quantitative link between changes in the residual state and the increasing selectivity observed across depth.

The fitted distributions predict competition along observed trajectories, leaving the origin and timing of residual updates unexplained. Studies of linguistic collapse examine geometry shaped by training \citep{wu2024linguistic}, while controlled sequence models relate residual geometry to predictive belief states \citep{shai2024belief}. Following endpoint populations and residual trajectories across training checkpoints could reveal how their organization develops together. The functional role of geometric competition also remains open. Competitor sets depend on the bank and metric, and their size reaches zero at the own endpoint by construction. Endpoint geometry captures only part of the variation in model outputs, and the observed turnover does not establish an explicit internal search process. Interventions near competitor-margin crossings could test whether changes in geometric preference affect token predictions.

Taken together, our empirical results and theoretical models support the geometric inference hypothesis that residual updates refine early endpoint preference into greater selectivity while individual comparisons can be revised. The endpoint population determines how residual movement translates into competition, providing a quantitative link between the geometry of final representations and the refinement of intermediate states.

\bibliography{paper}

@article{xu2026scale,
  title={Scale Determines Whether Language Models Organize Representation Geometry for Prediction},
  author={Xu, Weilun},
  journal={arXiv preprint arXiv:2605.17084},
  year={2026},
  url={https://arxiv.org/abs/2605.17084}
}

@misc{elhage2021framework,
  title        = {A Mathematical Framework for Transformer Circuits},
  author       = {Elhage, Nelson and Nanda, Neel and Olsson, Catherine and Henighan, Tom and Joseph, Nicholas and Mann, Ben and Askell, Amanda and Bai, Yuntao and Chen, Anna and Conerly, Tom and DasSarma, Nova and Drain, Dawn and Ganguli, Deep and Hatfield-Dodds, Zac and Hernandez, Danny and Jones, Andy and Kernion, Jackson and Lovitt, Liane and Ndousse, Kamal and Amodei, Dario and Brown, Tom and Clark, Jack and Kaplan, Jared and McCandlish, Sam and Olah, Chris},
  year         = {2021},
  howpublished = {Transformer Circuits Thread},
  url          = {https://transformer-circuits.pub/2021/framework/index.html}
}

@inproceedings{ethayarajh2019contextual,
  title     = {How Contextual are Contextualized Word Representations? {C}omparing the Geometry of {BERT}, {ELM}o, and {GPT}-2 Embeddings},
  author    = {Ethayarajh, Kawin},
  booktitle = {Proceedings of the 2019 Conference on Empirical Methods in Natural Language Processing and the 9th International Joint Conference on Natural Language Processing (EMNLP-IJCNLP)},
  pages     = {55--65},
  year      = {2019},
  doi       = {10.18653/v1/D19-1006}
}

@inproceedings{park2025geometry,
  title     = {The Geometry of Categorical and Hierarchical Concepts in Large Language Models},
  author    = {Park, Kiho and Choe, Yo Joong and Jiang, Yibo and Veitch, Victor},
  booktitle = {International Conference on Learning Representations},
  year      = {2025},
  url       = {https://openreview.net/forum?id=KXuYjuBzKo}
}

@article{lombardo2026geometric,
  title   = {A Geometric Perspective on Next-Token Prediction in Large Language Models: Three Emerging Phases},
  author  = {Lombardo, Gianfranco and Trimigno, Giuseppe and Cagnoni, Stefano},
  journal = {arXiv preprint arXiv:2605.09011},
  year    = {2026},
  url     = {https://arxiv.org/abs/2605.09011}
}

@article{guda2026stratification,
  title   = {Geometric and Behavioral Stratification in Transformer Residual Streams},
  author  = {Guda, Nelson},
  journal = {arXiv preprint arXiv:2608.12447},
  year    = {2026},
  url     = {https://arxiv.org/abs/2608.12447}
}

@book{mardia2000directional,
  title     = {Directional Statistics},
  author    = {Mardia, Kanti V. and Jupp, Peter E.},
  publisher = {Wiley},
  year      = {2000}
}

@article{banerjee2005vmf,
  title   = {Clustering on the Unit Hypersphere using von {M}ises--{F}isher Distributions},
  author  = {Banerjee, Arindam and Dhillon, Inderjit S. and Ghosh, Joydeep and Sra, Suvrit},
  journal = {Journal of Machine Learning Research},
  volume  = {6},
  pages   = {1345--1382},
  year    = {2005}
}

@article{belrose2023tuned,
  title={Eliciting Latent Predictions from Transformers with the Tuned Lens},
  author={Belrose, Nora and Ostrovsky, Igor and McKinney, Lev and Furman, Zach and Smith, Logan and Halawi, Danny and Biderman, Stella and Steinhardt, Jacob},
  journal={arXiv preprint arXiv:2303.08112},
  year={2023},
  url={https://arxiv.org/abs/2303.08112}
}

@article{ali2025entropy,
  title={{Entropy-Lens}: Uncovering Decision Strategies in {LLMs}},
  author={Ali, Riccardo and Caso, Francesco and Irwin, Christopher and Li\`o, Pietro},
  journal={arXiv preprint arXiv:2502.16570},
  year={2025},
  url={https://arxiv.org/abs/2502.16570}
}

@article{wang2013projected,
 title={Directional Data Analysis under the General Projected Normal Distribution},
 author={Wang, Fangpo and Gelfand, Alan E.},
 journal={Statistical Methodology}, volume={10}, number={1}, pages={113--127}, year={2013},
 doi={10.1016/j.stamet.2012.07.005}
}

@inproceedings{geva2022transformer,
  title={Transformer Feed-Forward Layers Build Predictions by Promoting Concepts in the Vocabulary Space},
  author={Geva, Mor and Caciularu, Avi and Wang, Kevin and Goldberg, Yoav},
  booktitle={Proceedings of the 2022 Conference on Empirical Methods in Natural Language Processing},
  pages={30--45},
  year={2022},
  publisher={Association for Computational Linguistics},
  doi={10.18653/v1/2022.emnlp-main.3},
  url={https://aclanthology.org/2022.emnlp-main.3/}
}

@inproceedings{valeriani2023geometry,
  title={The Geometry of Hidden Representations of Large Transformer Models},
  author={Valeriani, Lucrezia and Doimo, Diego and Cuturello, Francesca and Laio, Alessandro and Ansuini, Alessio and Cazzaniga, Alberto},
  booktitle={Advances in Neural Information Processing Systems},
  volume={36},
  year={2023},
  doi={10.52202/075280-2230},
  url={https://proceedings.neurips.cc/paper_files/paper/2023/hash/a0e66093d7168b40246af1cddc025daa-Abstract-Conference.html}
}

@article{jiang2026layerwise,
  title={Layer-wise Representation Dynamics: An Empirical Investigation Across Embedders and Base {LLMs}},
  author={Jiang, Jingzhou and Yang, Yi and Tam, Kar Yan},
  journal={arXiv preprint arXiv:2605.12714},
  year={2026},
  url={https://arxiv.org/abs/2605.12714}
}

@article{gu2026semrf,
  title={{SemRF}: A Semantic Reference Frame for Residual-Stream Dynamics in Language Models},
  author={Gu, Jian and Aleti, Aldeida and Chen, Chunyang and Zhang, Hongyu},
  journal={arXiv preprint arXiv:2606.32022},
  year={2026},
  url={https://arxiv.org/abs/2606.32022}
}

@inproceedings{wu2024linguistic,
  title={Linguistic Collapse: Neural Collapse in (Large) Language Models},
  author={Wu, Robert and Papyan, Vardan},
  booktitle={Advances in Neural Information Processing Systems},
  volume={37},
  year={2024},
  doi={10.52202/079017-4366},
  url={https://proceedings.neurips.cc/paper_files/paper/2024/hash/f88cc8930b47a45ec4733123bf3039b9-Abstract-Conference.html}
}

@inproceedings{shai2024belief,
  title={Transformers Represent Belief State Geometry in Their Residual Stream},
  author={Shai, Adam S. and Marzen, Sarah E. and Teixeira, Lucas and Gietelink Oldenziel, Alexander and Riechers, Paul M.},
  booktitle={Advances in Neural Information Processing Systems},
  volume={37},
  year={2024},
  doi={10.52202/079017-2387},
  url={https://proceedings.neurips.cc/paper_files/paper/2024/hash/8936fa1691764912d9519e1b5673ea66-Abstract-Conference.html}
}

@inproceedings{penedo2024fineweb,
  title={The {FineWeb} Datasets: Decanting the Web for the Finest Text Data at Scale},
  author={Penedo, Guilherme and Kydl{\'{i}}{\v{c}}ek, Hynek and Ben allal, Loubna and Lozhkov, Anton and Mitchell, Margaret and Raffel, Colin and Von Werra, Leandro and Wolf, Thomas},
  booktitle={Advances in Neural Information Processing Systems},
  volume={37},
  year={2024},
  doi={10.52202/079017-0970},
  url={https://proceedings.neurips.cc/paper_files/paper/2024/hash/370df50ccfdf8bde18f8f9c2d9151bda-Abstract-Datasets_and_Benchmarks_Track.html}
}

@article{gemmateam2024gemma,
  title={{Gemma}: Open Models Based on {Gemini} Research and Technology},
  author={{Gemma Team}},
  journal={arXiv preprint arXiv:2403.08295},
  year={2024},
  url={https://arxiv.org/abs/2403.08295}
}

@article{qwenteam2024qwen25,
  title={{Qwen2.5} Technical Report},
  author={{Qwen Team}},
  journal={arXiv preprint arXiv:2412.15115},
  year={2024},
  url={https://arxiv.org/abs/2412.15115}
}

@misc{mistralai2024mistralv03,
  title={{Mistral-7B-v0.3} Model Card},
  author={{Mistral AI}},
  year={2024},
  url={https://huggingface.co/mistralai/Mistral-7B-v0.3}
}

@misc{aimeta2024llama3card,
  title={{Llama 3} Model Card},
  author={{AI@Meta}},
  year={2024},
  url={https://github.com/meta-llama/llama3/blob/main/MODEL_CARD.md}
}
\bibliographystyle{plainnat}

\newpage
\appendix

\section{Geometric Derivations}
\label{app:theory}

Section~\ref{sec:theory} relates residual movement to endpoint preference,
competitor membership, and competitor count. We derive the margin updates
and straight-path result, then show how norm, alignment, and endpoint
geometry can produce different depth profiles for these measurements.
Throughout, the endpoint bank is fixed, comparisons are strict, and cosine
distance is used only for nonzero states and endpoints.

\subsection{Endpoint margins and straight-line convergence}
\label{app:marginproof}

The turnover measurements in Section~\ref{sec:turnover} require a criterion
for when an endpoint enters or leaves the competitor set. For an own
endpoint $y_i$ and an alternative $y_j$, expanding the squared Euclidean
margin in Eq.~\ref{eq:update_margins} gives
\begin{equation}
 m^E_{ij}(x)=2x^\top(y_i-y_j)+\norm{y_j}^2-\norm{y_i}^2.
 \label{eq:app_euclidean_margin}
\end{equation}
Since squaring preserves the ordering of nonnegative distances, $j$ is a
Euclidean competitor exactly when this margin is negative. For cosine
distance, write $u_k=y_k/\norm{y_k}$. Then
\begin{equation}
 d_C(x,y_j)-d_C(x,y_i)
 =\frac{x^\top(u_i-u_j)}{\norm{x}}
 =\frac{m^C_{ij}(x)}{\norm{x}}.
 \label{eq:app_cosine_margin}
\end{equation}
The positive denominator preserves the sign. Both margins are affine in
$x$, which gives the update identities in Eq.~\ref{eq:margin_changes}.
The directional margin determines cosine membership, although its
magnitude also includes the norm of the current state.

\textbf{Proposition (nested competitor sets on a straight segment).}
Let $x(t)=(1-t)x_0+t y_i$ for $0\leq t\leq1$. For any
$0\leq s\leq t\leq1$,
\begin{equation}
 \CC^{(E)}(t)\subseteq\CC^{(E)}(s),
 \qquad
 \CC^{(C)}(t)\subseteq\CC^{(C)}(s).
 \label{eq:app_nesting}
\end{equation}
The cosine statement applies at parameter values where the states are
nonzero. Endpoint norms need not be equal.

\textbf{Proof.}
For $s<1$, write $x(t)=(1-\lambda)x(s)+\lambda y_i$, where
$\lambda=(t-s)/(1-s)\in[0,1]$. Affinity gives, for either margin,
\begin{equation}
 m_{ij}(x(t))=(1-\lambda)m_{ij}(x(s))+\lambda m_{ij}(y_i).
\end{equation}
At the own endpoint,
\begin{equation}
 m^E_{ij}(y_i)=\norm{y_j-y_i}^2\geq0,
 \qquad
 m^C_{ij}(y_i)=\norm{y_i}(1-u_i^\top u_j)\geq0.
\end{equation}
If $j$ is outside the competitor set at $s$, its margin is nonnegative
there and remains nonnegative for all later $t$. The case $s=1$ is
immediate. $\square$

The proposition also holds under any nondecreasing reparameterization of
the segment. Thus a resolved entry in a fixed bank excludes monotone
straight-line convergence on that trajectory. It does not identify the
update mechanism or distinguish curvature from backtracking. Ties remain
outside the strict competitor set, including duplicate endpoints and,
under cosine distance, endpoints with the same direction. Entries close
to a tie must be interpreted at the numerical precision of the stored
states.

\subsection{Norm and alignment during a distance plateau}
\label{app:norm_alignment}

The Euclidean plateaus in Section~\ref{sec:turnover} can coexist with
directional progress because distance depends on both norm and alignment.
For $r=\norm{x}$, $R=\norm{y_i}$, $s=r/R$, and
$c=x^\top y_i/(rR)$, expansion gives
\begin{equation}
 \frac{\norm{x-y_i}^2}{R^2}=s^2+1-2sc.
\end{equation}
Along a differentiable path with fixed $y_i$,
\begin{equation}
 \frac{d}{dt}\frac{\norm{x-y_i}^2}{R^2}
 =2(s-c)s'-2sc'.
 \label{eq:app_distance_rate}
\end{equation}
An increase in alignment can therefore be offset by a change in norm.
In particular, setting $s=2c$ keeps the normalized Euclidean distance
equal to one while cosine distance $1-c$ decreases. The construction in
Section~\ref{sec:theory} uses $0<c<1/2$, so the intermediate-state norm
also remains below the endpoint norm. This identity establishes the
compatibility of the two profiles without specifying the residual
updates that produce them.

\subsection{High-dimensional competition with a shared component}
\label{app:shared_component}

The shared-axis model in Eq.~\ref{eq:shared_axis_model} isolates how a
small alignment advantage can yield early average preference and later
large reductions in competitor count. For a fixed query and own endpoint,
let $a=w^\top v_i$ and write
\begin{equation}
 c=\frac{x^\top y_i}{rR}
 =b\rho+\sqrt{(1-b^2)(1-\rho^2)}\,a.
\end{equation}
The score of an alternative has the same form with $a$ replaced by
$w^\top v_j$. The coefficient is positive under the assumptions
$|b|<1$ and $0\leq\rho<1$, so cosine competition is equivalent to
$w^\top v_j>a$. Equal endpoint norms also give
\begin{equation}
 \norm{x-y_j}^2-\norm{x-y_i}^2
 =2rR\sqrt{(1-b^2)(1-\rho^2)}(a-w^\top v_j),
\end{equation}
which yields the same competitor set under Euclidean distance.

For $V\sim\mathrm{Unif}(\mathbb S^{D-1})$, with $D\geq2$, rotational
symmetry gives the density of $Z=w^\top V$ as
\begin{equation}
 f_D(z)=\frac{\Gamma(D/2)}{\sqrt\pi\,\Gamma((D-1)/2)}
 (1-z^2)^{(D-3)/2},\qquad -1<z<1.
 \label{eq:app_spherical_density}
\end{equation}
One derivation writes $V=G/\norm{G}$ for a standard Gaussian vector
$G\in\RR^D$. Then $Z^2$ has a
$\operatorname{Beta}(1/2,(D-1)/2)$ distribution, and the two signs of
$Z$ are equally probable. The competitive mass is
$\overline F_D(a)=\int_a^1 f_D(z)\,dz$. For $M$ independent alternatives
drawn independently of the fixed query and own endpoint,
\begin{equation}
 N_{<}\mid x,y_i\sim\operatorname{Binomial}(M,\overline F_D(a)),
 \qquad
 \PP(N_{<}=0\mid x,y_i)=(1-\overline F_D(a))^M.
 \label{eq:app_binomial}
\end{equation}
Linearity of expectation gives Eq.~\ref{eq:spherical_tail_count} even
when alternatives are mutually dependent, provided their conditional
marginal score distributions remain $f_D$.

\textbf{Average preference and near-isolation.}
Since $\EE[Z]=0$, any $a>0$ makes the own endpoint closer on average in
cosine distance and in squared Euclidean distance. The latter statement
does not automatically extend to mean Euclidean distance, since taking
a square root changes the average. For small positive $a$,
\begin{equation}
 \overline F_D(a)=\frac12-\int_0^a f_D(z)\,dz,
\end{equation}
so average preference can coexist with a substantial competitor fraction.
For fixed $z$, the Gaussian representation implies
\begin{equation}
 \sqrt D\,Z\Rightarrow\mathcal N(0,1),
 \qquad
 \overline F_D(z/\sqrt D)\longrightarrow1-\Phi(z).
 \label{eq:app_normal_limit}
\end{equation}
Thus the central approximation is $\overline F_D(a)\approx
1-\Phi(\sqrt D\,a)$. It explains the alignment scale in the main text,
while quantitative estimates far into the tail require the exact
spherical distribution. A small probability of any remaining competitor
is ensured by $M\overline F_D(a)\ll1$, since
$\PP(N_{<}>0)\leq\EE[N_{<}]=M\overline F_D(a)$. This bound does not
require independence among alternatives.

\textbf{Rate of count reduction.}
For a differentiable alignment schedule $a(t)\in(-1,1)$,
\begin{equation}
 -\frac{d}{dt}\log\big(M\overline F_D(a(t))\big)
 =\frac{f_D(a(t))}{\overline F_D(a(t))}\,a'(t).
\end{equation}
This is Eq.~\ref{eq:contraction_rate}. The relative decrease depends on
both the alignment speed and the amount of endpoint mass near the
comparison boundary. With $b=\rho=0$ and $r/R=2a$, increasing $a$ in
$(0,1/2)$ reduces competition while keeping normalized Euclidean
distance equal to one.

\textbf{Population spread and own-endpoint alignment.}
The same model clarifies why whole-layer spread does not determine
competitor count. For independent endpoint directions $v_j,v_k$,
\begin{equation}
 \frac{\norm{y_j-y_k}^2}{R^2}
 =2(1-\rho^2)(1-v_j^\top v_k),
 \qquad
 \EE[v_j^\top v_k]=0,
 \quad
 \operatorname{Var}(v_j^\top v_k)=\frac1D.
\end{equation}
Pairwise squared distances therefore concentrate around
$2R^2(1-\rho^2)$ as $D$ grows. To vary own-endpoint alignment while
preserving the marginal directional population, draw $v_i$ uniformly and
draw $z_i$ uniformly on the unit sphere orthogonal to $v_i$ within
$m^\perp$. Define
\begin{equation}
 w_i(a)=a v_i+\sqrt{1-a^2}\,z_i.
\end{equation}
Rotational invariance makes $w_i(a)$ marginally uniform for each fixed
$a\in[-1,1]$, while $w_i(a)^\top v_i=a$. Taking $b=\rho$ and a common
state norm across contexts preserves the normalized population spread
as $a$ changes. Competitor counts can consequently decrease without a
decrease in that spread. This construction supports the separation of
population geometry and trajectory-specific refinement in
Appendix~\ref{app:ambient}.

\textbf{Unequal endpoint norms.}
The equal-radius assumption isolates the directional contribution. For
an alternative $Y=S U$ with $S>0$ and $\norm U=1$, the Euclidean
competitor condition becomes
\begin{equation}
 2r\left(S\frac{x^\top U}{r}-Rc\right)>S^2-R^2.
\end{equation}
Both norm and direction now affect membership. Cosine competition
remains invariant to positive radial rescaling. The fitted endpoint
models retain this radial variation through a common empirical norm
distribution (Appendix~\ref{sec:methods_endpoint_models}).

\section{A Shrinking Competitor Set Can Become Less Cohesive}
\label{app:cohesion}\label{app:cohesion_example}

Section~\ref{sec:turnover} distinguishes a reduction in competitor count
from an increase in similarity among the remaining endpoints. We give an
exact example in which a straight path produces nested competitor sets
whose mean pairwise distance increases after an exit. The example
establishes that shrinkage alone does not imply greater cohesion.

Let $e(\theta)=(\cos\theta,\sin\theta)$ and fix the unit-radius endpoints
\begin{equation}
 y_i=e(0),\qquad
 y_1=e(\pi/6),\qquad
 y_2=e(\pi/2),\qquad
 y_3=e(7\pi/12).
\end{equation}
Move from $x_0=e(\pi/3)$ toward $y_i$ along
$x(t)=(1-t)x_0+t y_i$. At $t=0$, the own dot-product score is $1/2$,
while the alternative scores are $\sqrt3/2$, $\sqrt3/2$, and
$\sqrt2/2$. All three alternatives compete. At $t=1/5$,
\begin{equation}
 x(1/5)=\left(\frac35,\frac{2\sqrt3}{5}\right),
 \qquad x(1/5)^\top y_i=\frac35,
\end{equation}
and the alternative scores are
\begin{equation}
 \frac{\sqrt3}{2},\qquad
 \frac{2\sqrt3}{5},\qquad
 \frac{9\sqrt2-\sqrt6}{20}.
\end{equation}
Only $y_1$ and $y_2$ remain competitive. Equal endpoint norms make
these rankings identical under Euclidean and cosine distance.

Measure within-set spread by the mean distance over unordered pairs of
distinct competitors. Initially, the mean cosine distance is
\begin{equation}
 \frac{(1-\cos(\pi/3))+(1-\cos(5\pi/12))+(1-\cos(\pi/12))}{3}
 =\frac{5-\sqrt6}{6}\approx0.425.
\end{equation}
After the exit it is $1-\cos(\pi/3)=1/2$. The mean Euclidean distance
increases from
\begin{equation}
 \frac{1+2\sin(5\pi/24)+2\sin(\pi/24)}{3}\approx0.826
\end{equation}
to $1$. Thus the count falls from three to two while the surviving set
becomes less cohesive under both metrics. The construction embeds in any
ambient dimension of at least two. It is a mathematical counterexample
to an implication between measurements, and does not establish a cohesion
trend in the observed transformer trajectories.

\section{Experimental Setup and Endpoint-Distribution Fitting}
\label{app:protocol}\label{sec:methods}

The geometric inference hypothesis is evaluated through comparisons with
a fixed population of final residual states. This section specifies the
data, observables, uncertainty conventions, and endpoint-model protocol
used in Sections~\ref{sec:experiments} and~\ref{sec:theory}. The
distribution-fitting experiment uses document splits to test whether
final-state geometry predicts competition along held-out trajectories.

\subsection{Language models, data, and residual trajectories}
\label{sec:models_data}

We use the following six base pretrained checkpoints.
\begin{center}
\small
\begin{tabular}{ll}
\toprule
Model & Checkpoint identifier \\
\midrule
Gemma-2B & \texttt{google/gemma-2b} \\
Qwen2.5-1.5B & \texttt{Qwen/Qwen2.5-1.5B} \\
Gemma-7B & \texttt{google/gemma-7b} \\
Mistral-7B & \texttt{mistralai/Mistral-7B-v0.3} \\
Qwen2.5-7B & \texttt{Qwen/Qwen2.5-7B} \\
Llama-3-8B & \texttt{meta-llama/Meta-Llama-3-8B} \\
\bottomrule
\end{tabular}
\end{center}
For each model, the primary sample contains $N=1{,}024$ non-overlapping
256-token contexts from FineWeb \texttt{sample-10BT}. Source-document
identities are retained for controls, splitting, and uncertainty estimates.

For context $i$, $x_{\ell,i}\in\RR^d$ is the residual state at the final
input position after block $\ell$, for $\ell=1,\ldots,L$, before final
normalization. The own endpoint is $y_i=x_{L,i}$, and
$\mathcal Y=\{y_1,\ldots,y_N\}$ is the empirical endpoint population.
There is no pre-block embedding state in the measured trajectory.
Primary measurements use the native residual coordinates and the
distances in Eq.~\ref{eq:metrics}, repeated here for reference:
\begin{equation}
 d_E(x,y)=\norm{x-y},\qquad
 d_C(x,y)=1-\frac{x^\top y}{\norm{x}\norm{y}}.
 \label{eq:methods_metrics}
\end{equation}
Euclidean trajectory-to-endpoint distances are displayed in units of the
query's endpoint norm $R_i=\norm{y_i}$,
\begin{equation}
 \widetilde d_E(x_{\ell,i},y_j)
 =\frac{d_E(x_{\ell,i},y_j)}{R_i}.
 \label{eq:methods_eucnorm}
\end{equation}
This common positive scale preserves every within-query comparison.
Endpoint-to-endpoint diagnostics use raw Euclidean distance unless an
additional normalization is stated.

\textbf{Numerical precision and uncertainty.}
Model inference uses reduced precision, followed by floating-point
geometric calculations. Primary trajectory-to-endpoint comparisons
evaluate all eligible pairs in GPU chunks.
Document-cluster bootstrap summaries use 1,000 resamples. For the
trajectory and population summaries, the bootstrap resamples document
weights on precomputed context statistics while keeping the reference
bank fixed. If $s_i$ is a context statistic and $w_i^{(b)}$ is the
multiplicity of its source document in resample $b$, then
\begin{equation}
 \widehat s^{(b)}=
 \frac{\sum_i w_i^{(b)}s_i}{\sum_i w_i^{(b)}}.
 \label{eq:legacyboot}
\end{equation}
These bands describe variation over query documents conditional on the
comparison bank. They do not jointly resample both endpoints of each
pair. Strict-count panels instead show one standard deviation across
contexts. Bank-size ranges and trajectory-prediction intervals are
specified in Appendices~\ref{app:banksize} and~\ref{app:prediction}.

\subsection{Preference and competitor fractions}
\label{sec:methods_observables}

The main analysis separates early average preference from selectivity
among individual alternatives. For an eligible bank $\mathcal B_i$ of
size $M_i$, define
\begin{align}
 D^{(m)}_{\rm own}(\ell,i)&=d_m(x_{\ell,i},y_i),\\
 D^{(m)}_{\rm other}(\ell,i)&=\frac{1}{M_i}
 \sum_{j\in\mathcal B_i}d_m(x_{\ell,i},y_j),\\
 \Delta^{(m)}_{\ell,i}&=D^{(m)}_{\rm other}(\ell,i)
 -D^{(m)}_{\rm own}(\ell,i).
 \label{eq:methods_delta}
\end{align}
For Euclidean trajectory summaries, both distances and their difference
are divided by $R_i$ before averaging across contexts. A positive margin
means that the own endpoint is closer than the average eligible
alternative. Individual competitors are defined by
\begin{align}
 \CC^{(m)}_{\ell,i}=\{j\in\mathcal B_i:
 d_m(x_{\ell,i},y_j)<d_m(x_{\ell,i},y_i)\},
 \label{eq:methods_competitors}\\
 A^{(m)}_{\ell,i}=\frac{|\CC^{(m)}_{\ell,i}|}{M_i}.
 \label{eq:methods_competition}
\end{align}
The primary bank contains all $j\ne i$, giving $M_i=N-1=1{,}023$.
Restricted banks use their own eligible sizes. The mean curve is the
arithmetic average of $A^{(m)}_{\ell,i}$ over the evaluated contexts.
The strict retrieval rank is $1+|\CC^{(m)}_{\ell,i}|$, with ties excluded.
All strict counts vanish at $x_{L,i}=y_i$ by construction. These
statistics measure specificity among contextual endpoints and are not
calibrated probabilities over output tokens.

\subsection{From endpoint counts to competitive mass}
\label{sec:methods_distributional}

The endpoint-model experiment tests the link between a comparison region
and the probability mass it contains. With $x_{\ell,i}$ and $y_i$ fixed,
an alternative $Y\sim P_Y$ competes with probabilities
\begin{align}
 q^E_{\ell,i}(P_Y)
 &=\PP_{Y\sim P_Y}\!\left[
 \norm{Y-x_{\ell,i}}<\norm{y_i-x_{\ell,i}}\right],
 \label{eq:methods_ballmass}\\
 q^C_{\ell,i}(P_U)
 &=\PP_{U\sim P_U}\!\left[
 u_{\ell,i}^{\top}U>u_{\ell,i}^{\top}u_i\right],
 \label{eq:methods_capmass}
\end{align}
where $u_{\ell,i}=x_{\ell,i}/\norm{x_{\ell,i}}$,
$u_i=y_i/\norm{y_i}$, and $U=Y/\norm{Y}$. The Euclidean region is an
open ball centered at the state and passing through the own endpoint.
The cosine region is an open cap of directions.

For the empirical distribution that assigns mass $1/M_i$ to each
eligible endpoint, these masses equal $A^{(m)}_{\ell,i}$ exactly.
For a sampled bank, they determine the expected fraction when its
alternatives have the stated distribution conditional on the fixed
query and own endpoint. The expected count is $M_iq^{(m)}_{\ell,i}$.
Independence among alternatives is needed for a binomial count law,
but not for this expectation. Fitting $P_Y$ only to final states then
tests whether endpoint geometry predicts competitive mass along
previously unseen residual trajectories.

\subsection{Common radial law and directional models}
\label{sec:methods_endpoint_models}

To isolate the contribution of directional structure, all five fitted
families share the same empirical distribution of endpoint norms. For
each fitting endpoint, write
\begin{equation}
 y_i=r_i u_i,\qquad r_i=\norm{y_i},\qquad u_i=y_i/r_i.
 \label{eq:radial_directional}
\end{equation}
Synthetic norms are sampled with replacement from the current fitting
split, independently of synthetic directions. The families therefore
share the same radial marginal and the same assumption of independence
between norm and direction. Differences among their predictions arise
from their directional models, although dependence between norm and
direction in the real population can remain unmatched.

\textbf{Uniform sphere.}
The isotropic baseline draws $U\sim\mathrm{Unif}(\mathbb S^{d-1})$
without fitted directional parameters.

\textbf{Uniform spherical cap.}
Estimate the global axis and a boundary cosine from the fitting directions:
\begin{align}
 \widehat\mu=\frac{\sum_{i\in{\rm fit}}u_i}
 {\norm{\sum_{i\in{\rm fit}}u_i}},
 \label{eq:global_axis}\\
 c_{\min}(q)=\operatorname{Quantile}_q
 \{\widehat\mu^\top u_i:i\in{\rm fit}\}.
 \label{eq:cap_quantile}
\end{align}
The model is uniform with respect to spherical surface area on
$\{u\in\mathbb S^{d-1}:\widehat\mu^\top u\geq c_{\min}(q)\}$.
Validation selects the quantile $q$, which controls the angular width.

\textbf{Single von Mises--Fisher distribution.}
The vMF density is proportional to $\exp(\kappa\mu^\top u)$ for a unit
mean direction $\mu$ and concentration $\kappa\geq0$. We use the axis
in Eq.~\ref{eq:global_axis} and estimate concentration from the resultant
length using the approximation of \citet{banerjee2005vmf}:
\begin{equation}
 \bar R=\left\|\frac{1}{n_{\rm fit}}\sum_{i\in{\rm fit}}u_i\right\|_2,
 \qquad
 \widehat\kappa=\frac{\bar R(d-\bar R^2)}{1-\bar R^2}.
 \label{eq:kappa_est}
\end{equation}
At $\kappa=0$ the distribution is uniform, and increasing $\kappa$
concentrates mass around the fitted axis.

\textbf{Mixture of vMF components.}
We partition unit endpoints into $C$ clusters by spherical $k$-means,
using multiple random initializations and retaining the smallest mean
cosine assignment loss. If $c_i$ is the assignment, let $n_c$ be the
cluster size and define
\begin{equation}
 \widehat\pi_c=\frac{n_c}{n_{\rm fit}},\qquad
 \widehat\mu_c=\frac{\sum_{i:c_i=c}u_i}{\norm{\sum_{i:c_i=c}u_i}},
 \qquad
 R_c=\left\|\frac{1}{n_c}\sum_{i:c_i=c}u_i\right\|_2.
\end{equation}
All components share a pooled concentration, obtained by substituting
$\bar R_{\rm mix}=\sum_c\widehat\pi_cR_c$ into
Eq.~\ref{eq:kappa_est}. The directional distribution is
\begin{equation}
 p(u)=\sum_{c=1}^{C}\widehat\pi_c\,
 \vMF(u;\widehat\mu_c,\widehat\kappa).
 \label{eq:vmf_mix}
\end{equation}
The shared concentration limits the number of component-specific
parameters as the mixture size increases.

\textbf{Low-rank projected normal.}
This family represents directional anisotropy by normalizing a Gaussian
sample \citep{wang2013projected}. Let $\bar u$ and $\widehat\Sigma$ be
the empirical mean and covariance of the fitting directions. Let
$v_1,\ldots,v_k$ be the leading covariance eigenvectors, with eigenvalues
$\lambda_1,\ldots,\lambda_k$. For $k<d$, the remaining variance is
represented by
\begin{equation}
 \sigma^2=\frac{\operatorname{tr}(\widehat\Sigma)
 -\sum_{j=1}^{k}\lambda_j}{d-k}.
 \label{eq:pn_sigma}
\end{equation}
The latent Gaussian and projected direction are
\begin{align}
 z&=\bar u+\sum_{j=1}^{k}\sqrt{(\lambda_j-\sigma^2)_+}\,a_jv_j
 +\sigma\epsilon,
 \quad a_j\sim\mathcal N(0,1),\quad\epsilon\sim\mathcal N(0,I_d),
 \label{eq:pn_latent}\\
 U&=z/\norm{z},
 \label{eq:pn_project}
\end{align}
with independent Gaussian draws and $(t)_+=\max(t,0)$. Subtracting the
noise variance in the retained directions prevents the isotropic term
from adding that variance twice. This is a moment-based approximation
to the observed unit endpoints. Projection changes the Gaussian moments,
so the generated directions need not have mean $\bar u$ or covariance
$\widehat\Sigma$. The fitted axes describe statistical variation without
an established semantic interpretation.

\begin{table}[t]
\centering
\small
\begin{tabular}{lll}
\toprule
Family & Directional parameters & Selected complexity \\
\midrule
Sphere & none & none \\
Cap & $\widehat\mu,c_{\min}(q)$ & boundary quantile $q$ \\
vMF & $\widehat\mu,\widehat\kappa$ & none \\
vMF mixture & $\{\widehat\pi_c,\widehat\mu_c\}_{c=1}^C,\widehat\kappa$
 & components $C$ \\
Projected normal & $\bar u,\{v_j,\lambda_j\}_{j=1}^k,\sigma$
 & rank $k$ \\
\bottomrule
\end{tabular}
\caption{\textbf{Directional endpoint models.} Parameters are fitted to unit
endpoints, and all families share the empirical norm distribution of the
fitting split. Complexity is selected using validation endpoints.}
\label{tab:fitio}
\end{table}

\subsection{Document-level fitting and model selection}
\label{sec:methods_fit}

To evaluate transfer to unseen contexts, we split source documents
approximately 60/20/20 into training, validation, and test sets. Contexts
from the same document remain in one partition. During model selection,
parameters and the radial law use training endpoints only. After
validation selects a configuration within each family, we refit it on
the union of training and validation endpoints. Intermediate residual
states, output probabilities, output-token labels, and competitor
curves are excluded from fitting and selection.

\begin{figure}[t]
\centering
\fbox{\begin{minipage}{0.94\linewidth}
\small
\textbf{Input:} Final residual endpoints, document IDs, an endpoint-model
family $\MM$, and a candidate complexity grid $\mathcal H_{\MM}$.
\begin{enumerate}
\item Split documents into training, validation, and test partitions.
\item For each candidate $h\in\mathcal H_{\MM}$, fit the directional model
and empirical norm distribution using training endpoints. Generate 12
synthetic samples, each matching the number of validation endpoints.
\item Average the diagnostic score in Eq.~\ref{eq:validation_score}
over these samples and select the lowest-scoring configuration. Extend
an upper-boundary search where feasible, using validation data only.
\item Refit the selected configuration and norm distribution on training
plus validation endpoints. Evaluate the endpoint diagnostics on test data.
\item Draw fixed alternative banks from the refitted model and evaluate
competition along the observed test trajectories.
\end{enumerate}
\textbf{Output:} A validation-selected endpoint distribution and its
held-out endpoint-fit and trajectory-prediction scores.
\end{minipage}}
\caption{\textbf{Fitting endpoint geometry before evaluating trajectories.}
The protocol tests whether final-state structure predicts competitive
mass without fitting the observed competitor curves.}
\label{alg:endpoint_fit}
\end{figure}

\subsection{Matched endpoint-distribution diagnostics}
\label{sec:methods_validation}

The validation score compares broad directional geometry, radial
spacing, and local neighborhoods. For each real or synthetic endpoint
sample, we form four empirical distributions:
\begin{equation}
 \{u_i^\top u_j:i<j\},\quad
 \{\norm{y_i-y_j}:i<j\},\quad
 \{\min_{j\ne i}\norm{y_i-y_j}:i\},\quad
 \{\widehat\mu^\top u_i:i\}.
\end{equation}
They measure pairwise cosine similarity, pairwise Euclidean distance,
Euclidean nearest-neighbor distance, and cosine similarity to the fitted
global axis. The same axis, estimated from the fitting partition, is
used for both real and synthetic samples. Matching sample sizes is
particularly relevant for nearest-neighbor spacing.

Let $KS_k$ be the two-sample Kolmogorov--Smirnov distance for diagnostic
$k$. The selection score is
\begin{equation}
 S_{\rm val}=\frac14\left(
 KS_{\rm pair\text{-}cos}+KS_{\rm pair\text{-}E}
 +KS_{\rm NN\text{-}E}+KS_{\rm opening\text{-}cos}\right).
 \label{eq:validation_score}
\end{equation}
Lower scores indicate closer agreement. Pairwise diagnostics use at most
20,000 unordered pairs. Each candidate is evaluated with 12 synthetic
replicates, and its scores are averaged before selection. The KS values
are distributional discrepancies, without an independence assumption
for overlapping pairs or an interpretation as hypothesis-test $p$-values.

The initial complexity grids are
\begin{align}
 q_{\rm cap}&\in\{0.01,0.05,0.10,0.20,0.30,0.50,0.75\},\\
 C_{\rm mix}&\in\{2,4,8,16,32\},\\
 k_{\rm PN}&\in\{1,4,16,32,64,128,256\}.
\end{align}
Upper-boundary searches are extended to $C=64$ and $k=512$ where the
sample size permits. Mixture fits require sufficient training endpoints
per component and use multiple spherical-$k$-means initializations.
The sphere and single-vMF families have no discrete complexity parameter.
Appendix~\ref{app:models} reports the resulting selection curves and
held-out comparisons.

\subsection{Predicting competitor fractions from fitted endpoints}
\label{sec:methods_prediction}

The prediction experiment asks whether the selected endpoint distributions
assign the correct competitive mass to regions encountered by real
trajectories. For each held-out context, the state $x_{\ell,i}$ and own
endpoint $y_i$ remain fixed. For replicate $s$, draw
$\widetilde{\mathcal Y}^{(s)}=\{\widetilde Y^{(s)}_1,\ldots,
\widetilde Y^{(s)}_B\}$ from the refitted family $\MM$, and compute
\begin{align}
 \widehat A^{(m,s)}_{\MM}(\ell,i)
 &=\frac1B\sum_{b=1}^B\ind\!\left[
 d_m(x_{\ell,i},\widetilde Y^{(s)}_b)<d_m(x_{\ell,i},y_i)\right],
 \label{eq:methods_predA_query}\\
 \widehat A^{(m)}_{\MM}(\ell)
 &=\frac{1}{8N_{\rm test}}\sum_{s=1}^{8}\sum_{i\in{\rm test}}
 \widehat A^{(m,s)}_{\MM}(\ell,i).
 \label{eq:methods_predA}
\end{align}
We use $B=256$ alternatives in each of eight independent banks. Each
bank is shared across test queries and held fixed across layers.
The empirical target $A^{(m)}(\ell)$ uses all test endpoints except the
query's own, giving $N_{\rm test}-1$ alternatives per query.

The real-endpoint reference uses equally sized banks sampled from
training endpoints. Its documents are disjoint from the test queries,
as are the training-plus-validation documents used by the synthetic
models. The empirical test bank can include other contexts from the
query's document. Thus the three comparisons differ in their sampling
pools. The real-bank reference measures finite-sample transfer across
documents and is not an error lower bound, since a fitted distribution
can reduce sampling variation through smoothing.

Prediction accuracy uses the same mean curve and offset as
Eq.~\ref{eq:logerr}:
\begin{equation}
 \operatorname{err}_{\log}
 =\frac{1}{L-1}\sum_{\ell=1}^{L-1}
 \left|\log_{10}\!\left(\widehat A_{\MM}(\ell)+\varepsilon\right)
 -\log_{10}\!\left(A(\ell)+\varepsilon\right)\right|,
 \qquad\varepsilon=10^{-3}.
 \label{eq:methods_logerr}
\end{equation}
The average across banks and queries is taken before applying the
logarithm. The final layer is excluded because strict competitor
fractions are zero there. The offset keeps the error defined at zero
and limits sensitivity to differences far below $10^{-3}$.

Uncertainty estimates use 1,000 document-cluster bootstrap resamples of
test queries. Each resample recomputes the context-weighted mean curves
and their log-scale error, conditional on the fitted distributions and
sampled comparison banks. Comparisons between families use the same
resampled document weights. These intervals do not include uncertainty
from refitting the endpoint models. The endpoint fit and subsequent
trajectory evaluation are summarized by
\begin{equation}
 \mathcal Y_{\rm train}\cup\mathcal Y_{\rm val}
 \xrightarrow{\text{refit}}\widehat P_{\MM}
 \xrightarrow{\text{sample}}\widetilde{\mathcal Y}
 \xrightarrow[\text{fixed }y_i]{\text{fixed }x_{\ell,i}}
 \widehat A_{\MM}(\ell).
 \label{eq:fit_pipeline}
\end{equation}

\section{Euclidean Trajectories and Competitor Counts}
\label{app:metriccompanions}

\begin{figure}[t]
\centering
\includegraphics[width=0.94\linewidth]{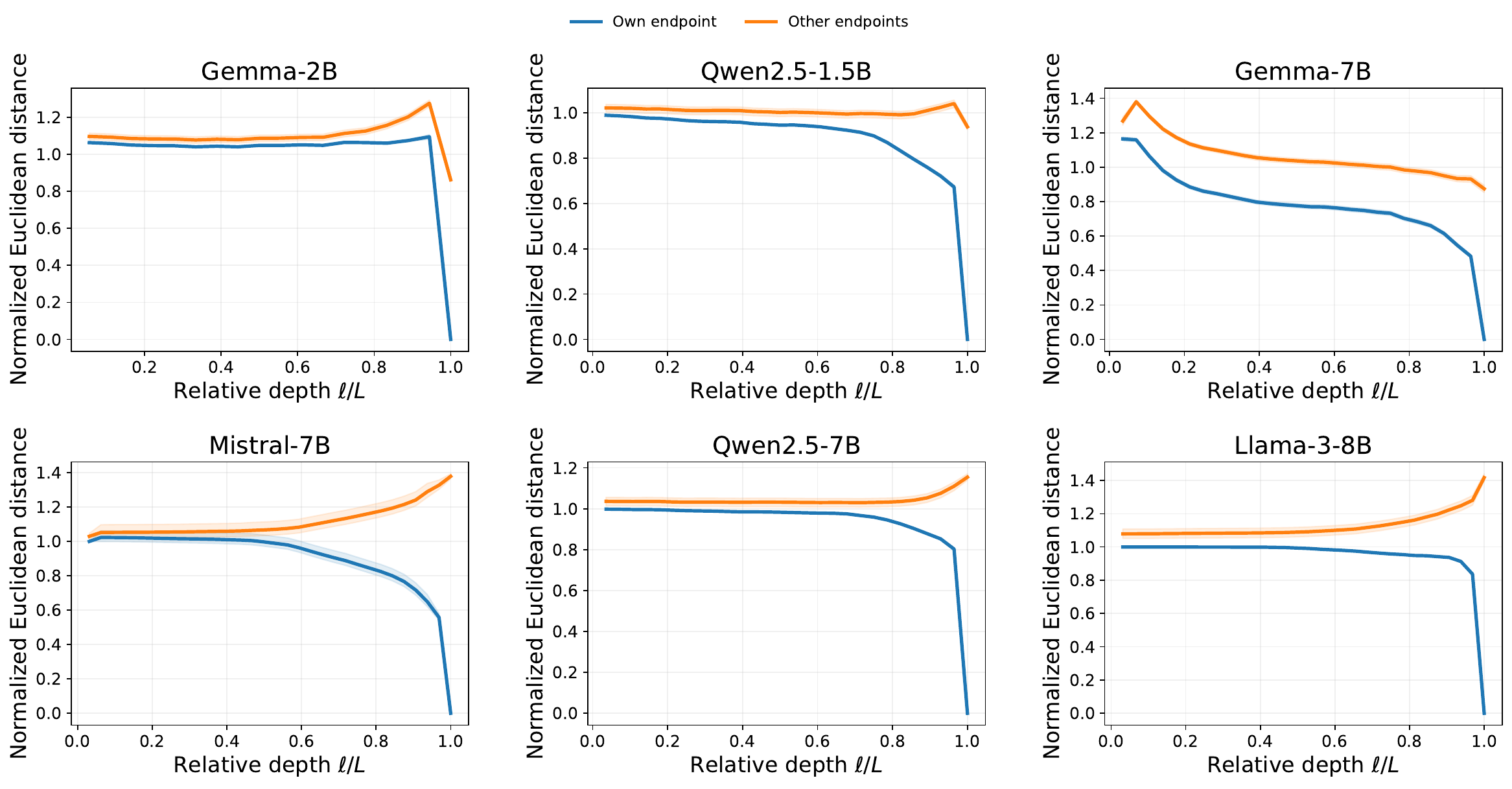}\par
{\small\textbf{(a) Distance to the own endpoint and mean alternative}}\par\smallskip
\includegraphics[width=0.94\linewidth]{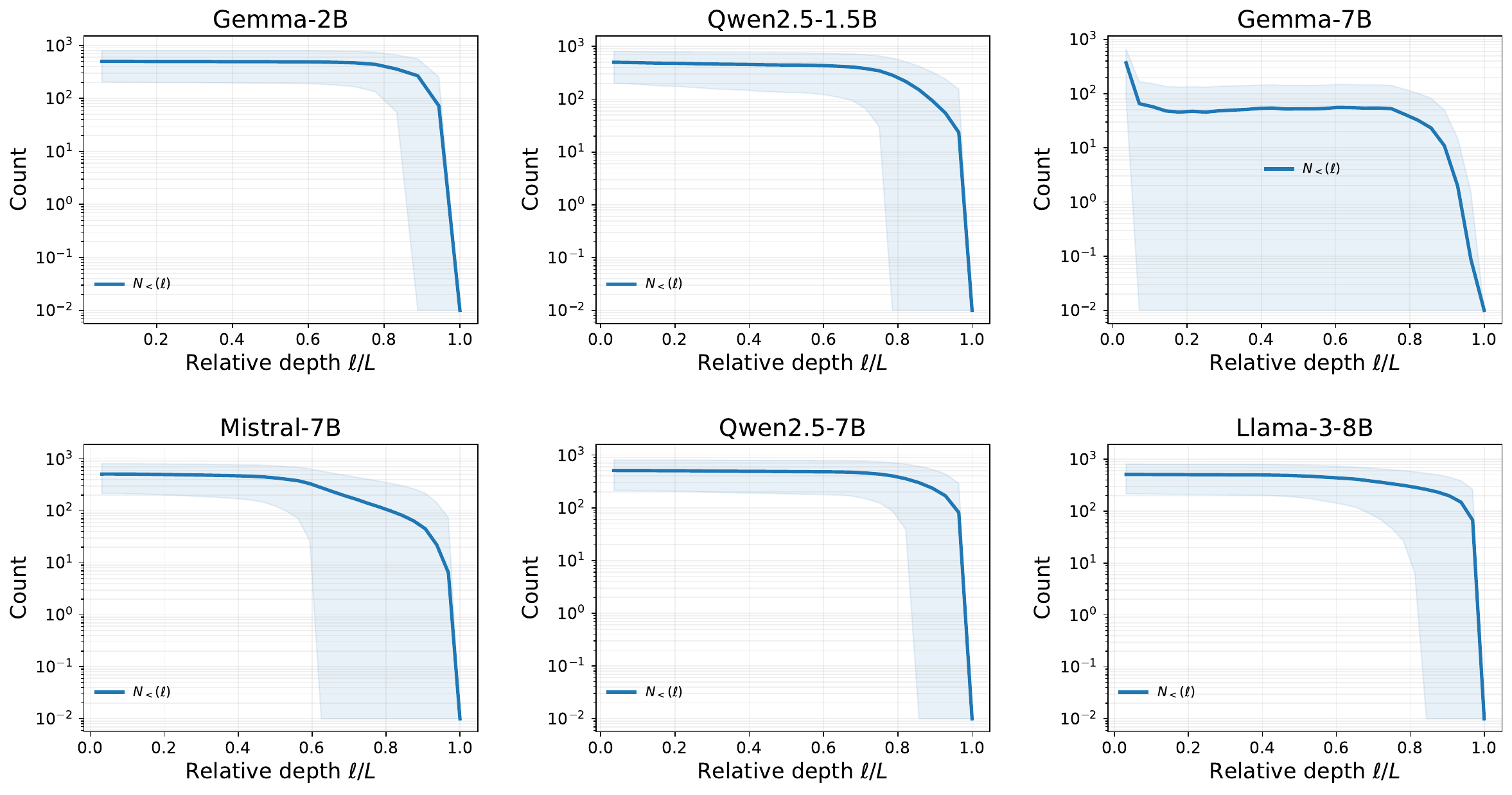}\par
{\small\textbf{(b) Number of strict competitors}}
\caption{\textbf{Euclidean preference and competitor counts.}
Distances in (a) are divided by the query's endpoint norm. Panel (b)
shows mean strict counts with one standard deviation across contexts.
The final count is zero by construction and is displayed at $10^{-2}$
on the logarithmic axis.}
\label{fig:trajectory_euc}
\end{figure}

The cosine results in Sections~\ref{sec:exp_preference} and~\ref{sec:turnover}
describe directional specificity. Their Euclidean counterparts show how
endpoint norm changes the timing of preference and competitor shrinkage.
Figure~\ref{fig:trajectory_euc} reports distance to the own endpoint,
mean distance to alternatives, and strict competitor count.

The mean distance gap is small over much of depth in Gemma-2B,
Qwen2.5-7B, and Llama-3-8B. It is larger in Gemma-7B and widens during
intermediate or later layers in the other models. In five models, counts
remain in the hundreds across substantial portions of depth before
falling sharply. Gemma-7B reaches a smaller competitor set earlier.
Together with Figure~\ref{fig:trajectory}, these results show that early
average preference can coexist with many competitors and that the timing
of increasing specificity depends on both architecture and metric.

\section{Early Own-Endpoint Preference}
\label{app:separation}

Section~\ref{sec:exp_preference} reports that average own-endpoint
preference is detectable before the competitor set becomes small.
To assess this average at the document level, we first average the
context margins $\Delta_{\ell,i}$ within each source document and then
apply a one-sided one-sample $t$-test for a positive mean document margin.
Euclidean margins use the query normalization in
Eq.~\ref{eq:methods_eucnorm}.

\begin{figure}[t]
\centering
\begin{minipage}[t]{0.49\linewidth}\centering\includegraphics[width=\linewidth]{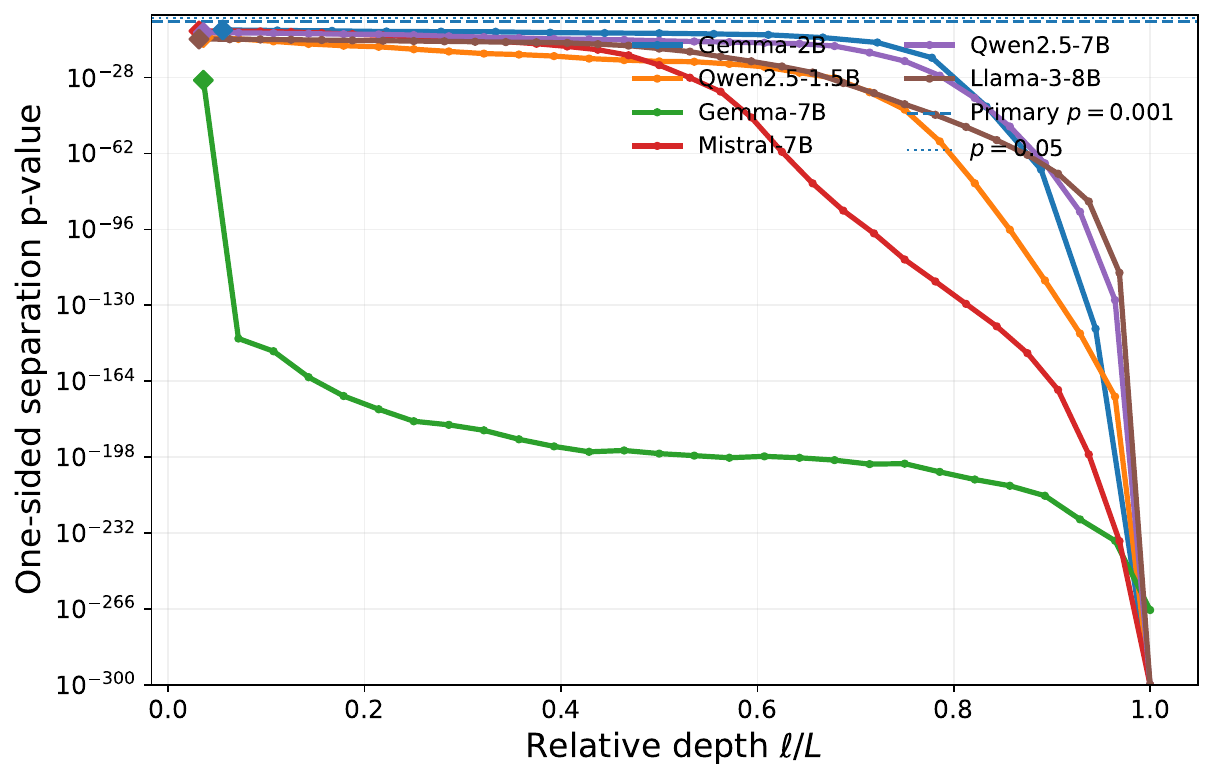}\end{minipage}\hfill
\begin{minipage}[t]{0.49\linewidth}\centering\includegraphics[width=\linewidth]{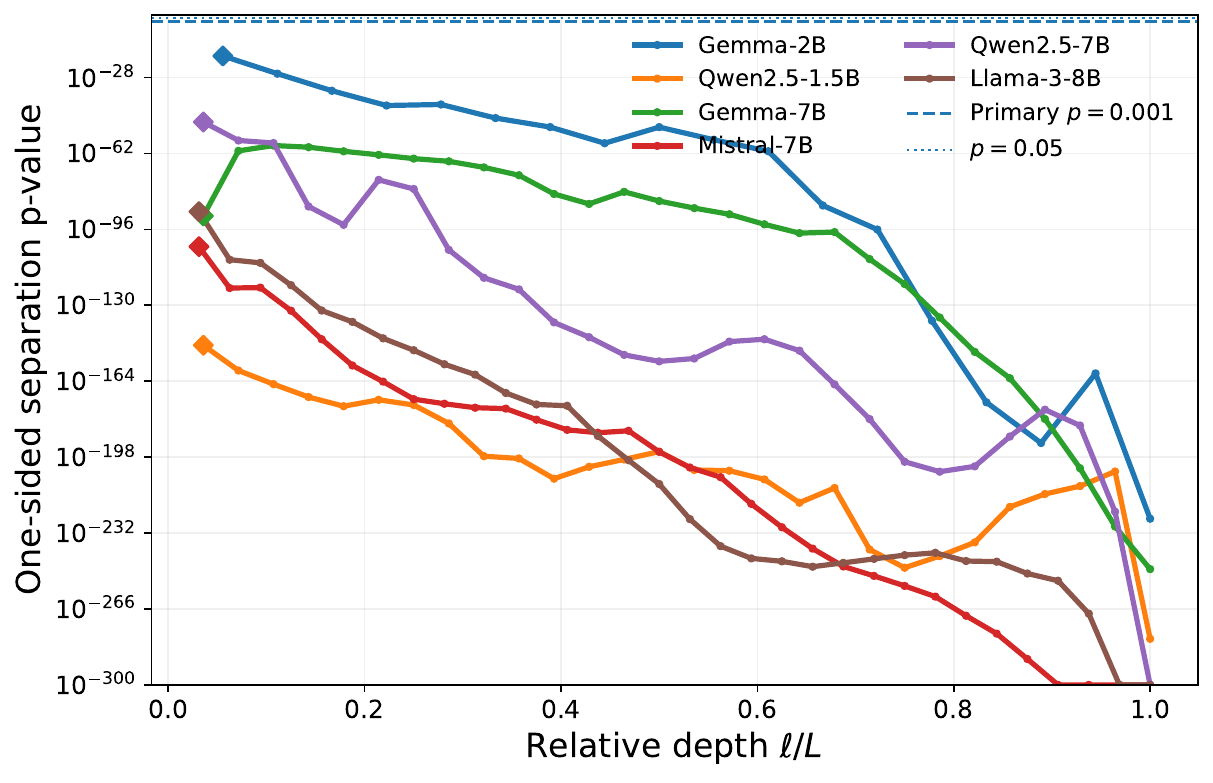}\end{minipage}
\caption{\textbf{Document-level tests of average own-endpoint preference.}
One-sided tests under Euclidean (left) and cosine (right) distance.
The working threshold is $p<10^{-3}$ without correction across layers.
Values below the plotting floor are shown at $10^{-300}$.}
\label{fig:pvalues}
\end{figure}

All six models meet the working $p<10^{-3}$ threshold at the first
measured post-block layer under both metrics (Figure~\ref{fig:pvalues}).
This result concerns the mean across documents and does not imply
positive preference for every context or locate a unique transition in
computation. Document means receive equal weight in the test, whereas
the trajectory curves average contexts equally. These conventions differ
when documents contribute different numbers of contexts. The layerwise
$p$-values are unadjusted for multiple testing.

\section{Input-Token and Document Controls}
\label{app:tokencontrols}

The interpretation of early preference in Section~\ref{sec:turnover}
depends on how much it reflects shared document content or the final
input token. We examine these contributions by restricting the
alternative bank and removing the input-token embedding direction.
Each control retains the comparison with the query's own endpoint.

\subsection{Excluding alternatives from the query's document}

The primary 1,024-context sample contains 333--361 source documents,
depending on the model and tokenizer preprocessing. Although 876--898
contexts have another sampled context from the same document, only
5.4--6.1 of the 1,023 alternatives per query share that document on
average. We exclude these alternatives and divide counts by the
remaining eligible bank size.

\begin{figure}[t]
\centering
{\small\textcolor[rgb]{0.122,0.467,0.706}{\rule{1em}{1.2pt}} full bank
\qquad\textcolor[rgb]{1.000,0.498,0.055}{\rule{1em}{1.2pt}} excluding the query's document}\par\smallskip
\includegraphics[width=0.94\linewidth,trim=0 0 0 40pt,clip]{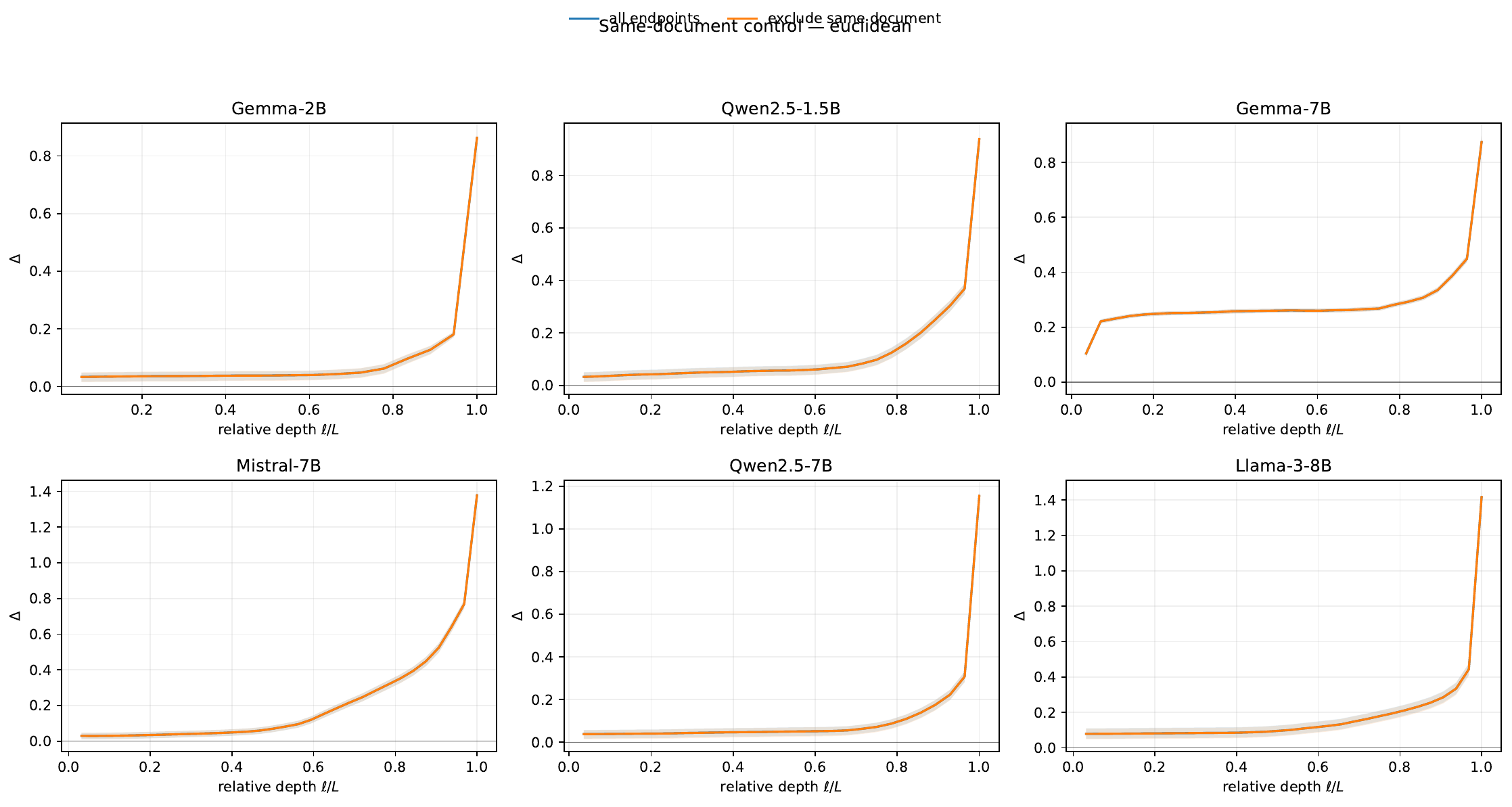}\par
{\small\textbf{(a) Euclidean distance}}\par\smallskip
\includegraphics[width=0.94\linewidth,trim=0 0 0 40pt,clip]{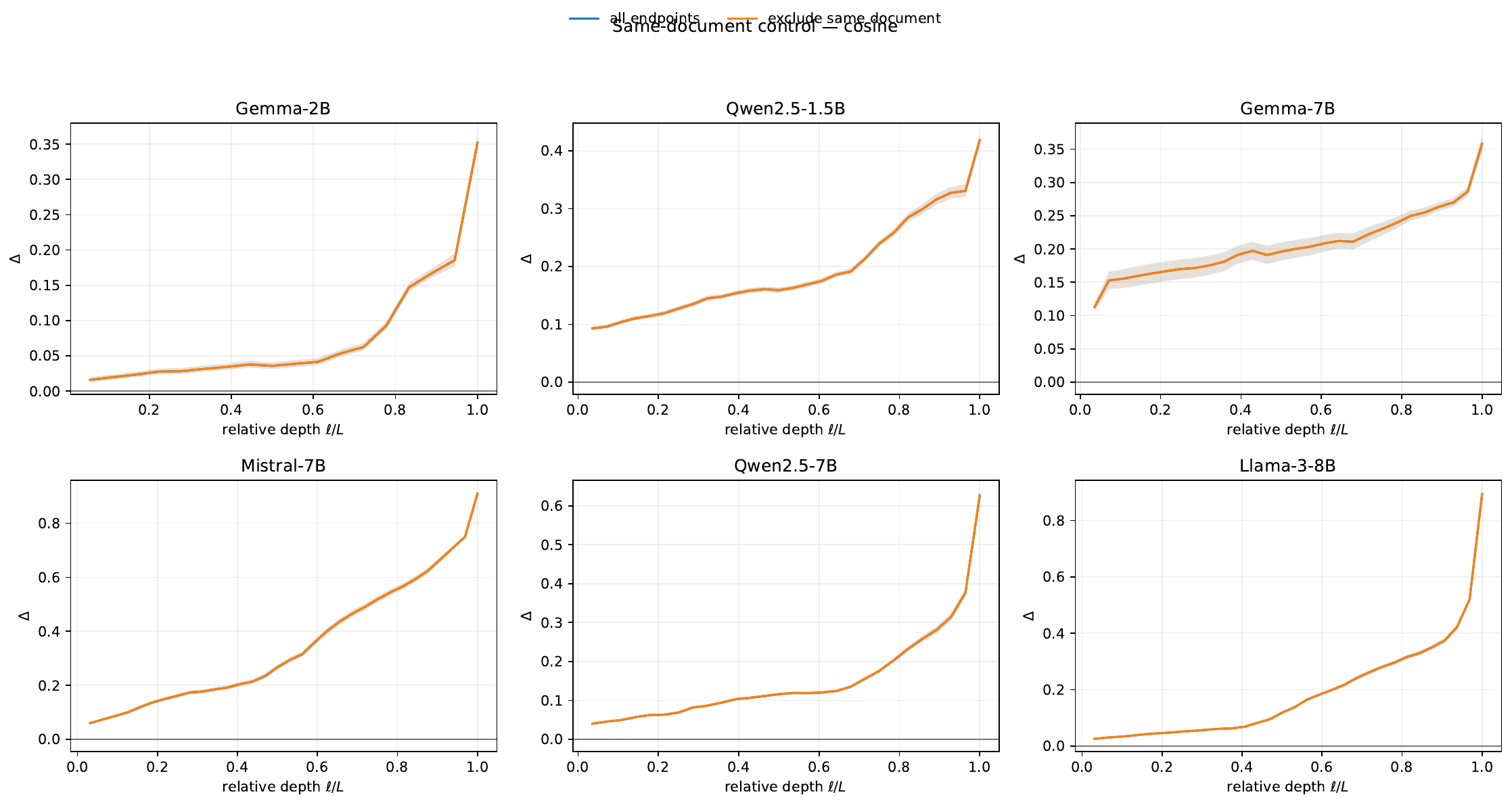}\par
{\small\textbf{(b) Cosine distance}}
\caption{\textbf{Average preference after excluding same-document alternatives.}
Euclidean (a) and cosine (b) margins with the full bank and the
bank excluding endpoints from the query's document.}
\label{fig:documentcontrol}
\end{figure}

The cosine preference and competitor-fraction curves change little, and
first-layer average preference remains positive in every model.
Euclidean effect sizes also change little, although the $p<10^{-3}$
threshold crossing moves by a few layers when early margins are small.
Figure~\ref{fig:documentcontrol} compares the preference margins.
These results indicate that the early average preference is not
accounted for by same-document alternatives in the primary bank.

\subsection{Matching the final input token}

A separate sample contains 2,000 contexts per model, with 40 contexts
for each of 50 frequent final input tokens, drawn from distinct source
documents. For each query, the same-token bank contains the other 39
endpoints with that final input token. A size-matched different-token
bank contains 39 alternatives. Both banks remain fixed across layers.

\begin{figure}[t]
\centering
\includegraphics[width=\linewidth]{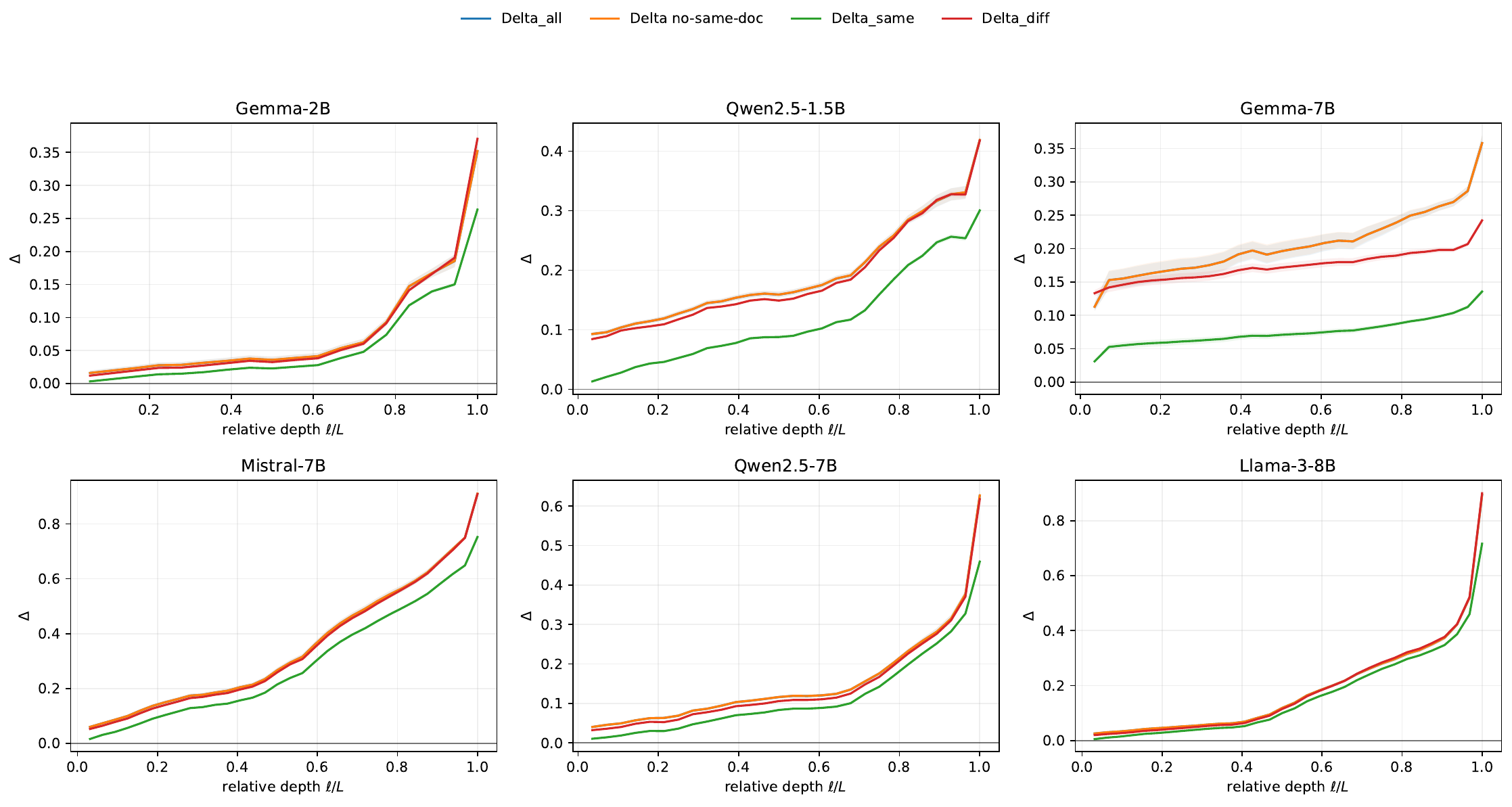}
\caption{\textbf{Cosine average preference with input-token controls.}
Mean margins for the original bank, the bank excluding the query's
document, 39 same-input-token alternatives, and 39 different-token
alternatives. The matched token banks use the separate 2,000-context sample.}
\label{fig:tokencontrol}
\end{figure}

\begin{figure}[t]
\centering
\includegraphics[width=\linewidth]{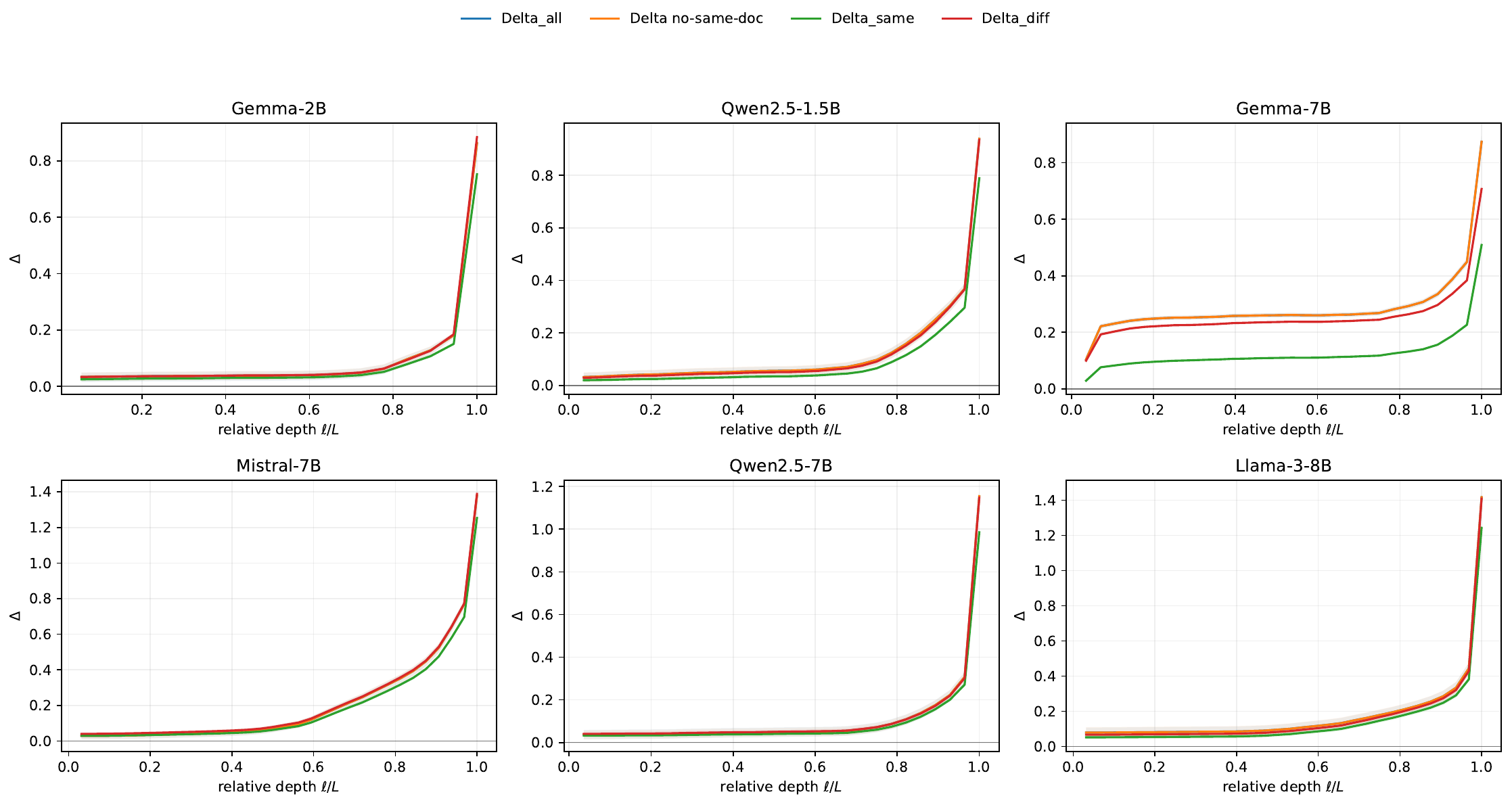}
\caption{\textbf{Euclidean average preference with input-token controls.}
The own endpoint is closer than the average same-token alternative from
the first measured layer in all six models. Curves show preference
margins for the same four bank conditions as Figure~\ref{fig:tokencontrol}.}
\label{fig:tokencontrol_euc}
\end{figure}

Average preference is positive from the first measured layer in all six
models under both metrics (Figures~\ref{fig:tokencontrol}
and~\ref{fig:tokencontrol_euc}). Same-token margins are smaller,
and more same-token alternatives remain competitive. At the first layer,
the cosine fractions for Qwen2.5-1.5B are $A_{\rm same}=0.364$ and
$A_{\rm diff}=0.058$. For Gemma-7B they are $0.253$ and $0.059$.
Input-token identity therefore contributes to early similarity, while
preference for the own endpoint persists among contexts sharing that token.

\subsection{Removing the input-token embedding direction}

\begin{figure}[t]
\centering
\includegraphics[width=\linewidth]{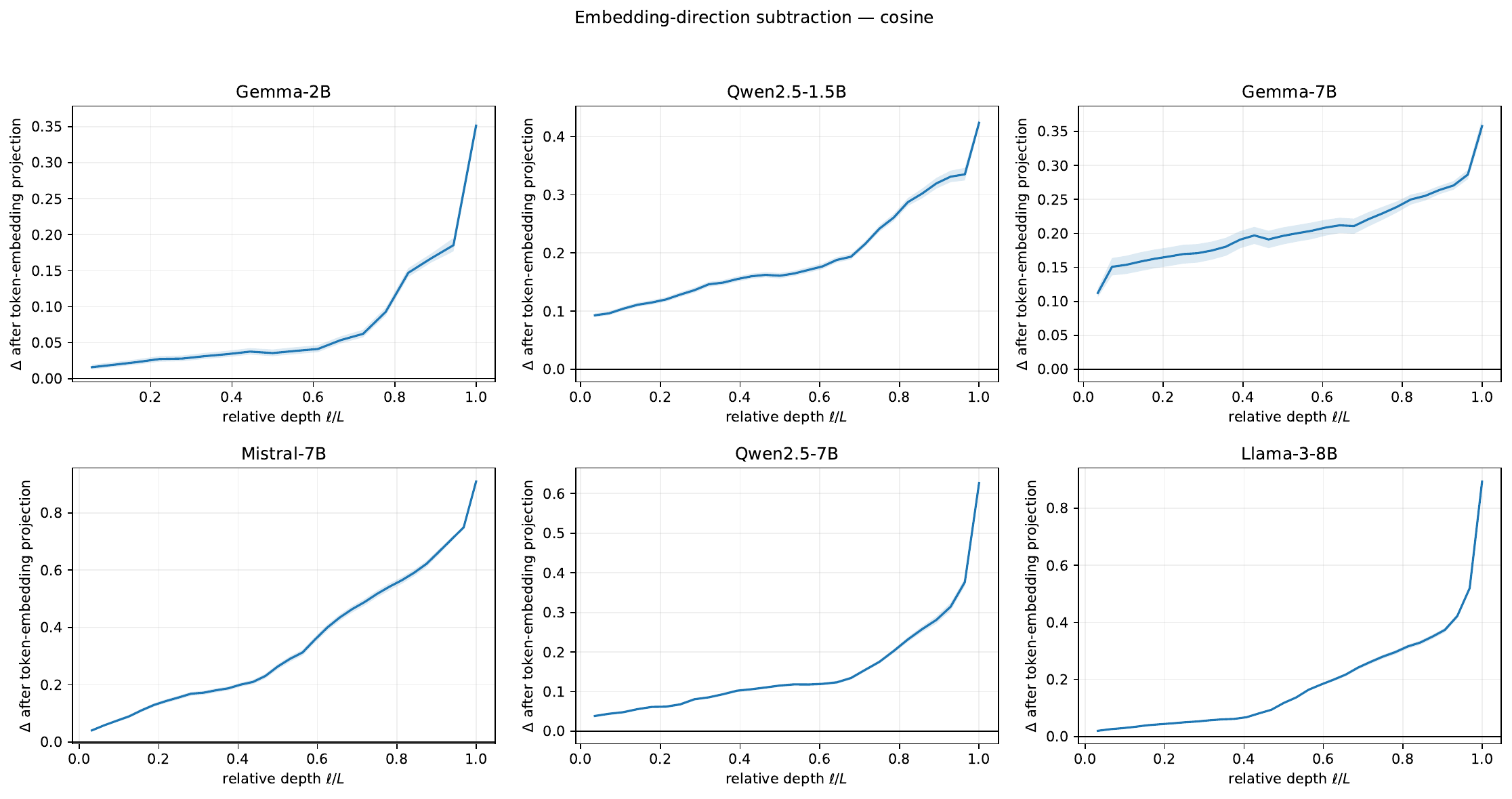}
\caption{\textbf{Cosine average preference after embedding-direction removal.}
The own-endpoint margin remains positive from the first measured layer
in all six models.}
\label{fig:embeddingproj_cos}
\end{figure}

\begin{figure}[t]
\centering
\includegraphics[width=\linewidth]{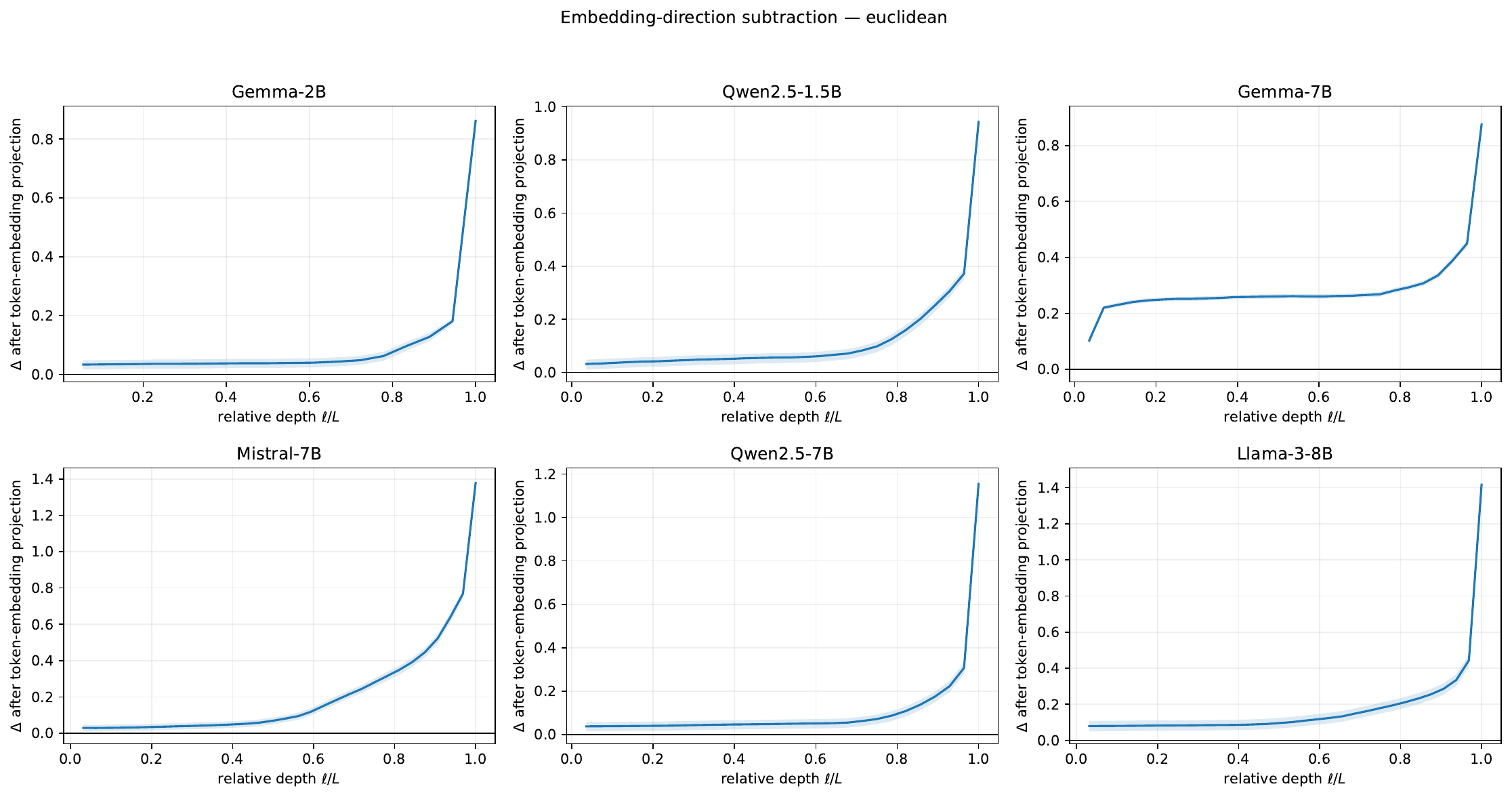}
\caption{\textbf{Euclidean average preference after embedding-direction removal.}
The projection changes the early margin more strongly than under cosine
distance, including delayed positive mean preference in both Qwen models
and Mistral-7B.}
\label{fig:embeddingproj_euc}
\end{figure}

The matching control retains the final input token, so we also examine
preference after a linear projection removes its embedding direction.
For the query trajectory, let $e_i$ be the normalized embedding of its
final input token. The transformed state and own endpoint are
\begin{equation}
 x'_{\ell,i}=x_{\ell,i}-(x_{\ell,i}^{\top}e_i)e_i,
 \qquad
 y'_i=y_i-(y_i^{\top}e_i)e_i.
 \label{eq:tokenproj}
\end{equation}
First-layer cosine preference remains positive in all six models, with
the mean margin ranging from $0.0159$ in Gemma-2B to $0.1124$ in
Gemma-7B (Figure~\ref{fig:embeddingproj_cos}). The Euclidean result is
more sensitive to this projection. The earliest layer with positive
mean preference moves to 5 in Qwen2.5-1.5B, 2 in Mistral-7B, and 13 in
Qwen2.5-7B (Figure~\ref{fig:embeddingproj_euc}). The first-layer
Euclidean competitor fraction is near one half in five models.

Together with the matched-bank result, the surviving cosine preference
supports a contribution from the broader context. The projection removes
one linear direction, so token information represented elsewhere and
information carried through residual connections can still contribute.

\section{Competitor Membership and Euclidean Turnover}
\label{app:turnover}

The changing membership in Section~\ref{sec:turnover} distinguishes
increasing specificity from successive removal of a fixed set of
alternatives. We define the composition statistics and compare the
Euclidean result with the cosine measurements in Figure~\ref{fig:composition}.
For adjacent measured layers, omit the metric superscript and let
\begin{align}
 J_{\ell,i}&=\frac{|\CC_{\ell-1,i}\cap\CC_{\ell,i}|}
 {|\CC_{\ell-1,i}\cup\CC_{\ell,i}|},\nonumber\\
 E_{\ell,i}&=\CC_{\ell,i}\setminus\CC_{\ell-1,i},
 &X_{\ell,i}&=\CC_{\ell-1,i}\setminus\CC_{\ell,i},\nonumber\\
 e_{\ell,i}&=\frac{|E_{\ell,i}|}{\max(|\CC_{\ell,i}|,1)},
 &f_{\ell,i}&=\frac{|X_{\ell,i}|}{\max(|\CC_{\ell-1,i}|,1)}.
 \label{eq:composition_app}
\end{align}
The entry fraction uses the current set size, and the exit fraction
uses the previous set size. Statistics are computed per context before
averaging. Both empty sets give $J=1$ and zero entry and exit fractions.
A transition from a nonempty to an empty set gives $J=0$, $e=0$, and
$f=1$. The reverse gives $J=0$, $e=1$, and $f=0$. The first measured
layer has no preceding transition.

High late-depth mean overlap can therefore include empty-to-empty
transitions. Entry includes both first appearances and returns of earlier
competitors. Also, entry and exit fractions have different denominators,
so their difference is not the change in competitor fraction. For a fixed
eligible bank, the count change is exactly
\begin{equation}
 |\CC_{\ell,i}|-|\CC_{\ell-1,i}|=|E_{\ell,i}|-|X_{\ell,i}|.
\end{equation}
This identity explains how entries can occur over depths where counts fall.

\begin{figure}[t]
\centering
\includegraphics[width=\linewidth]{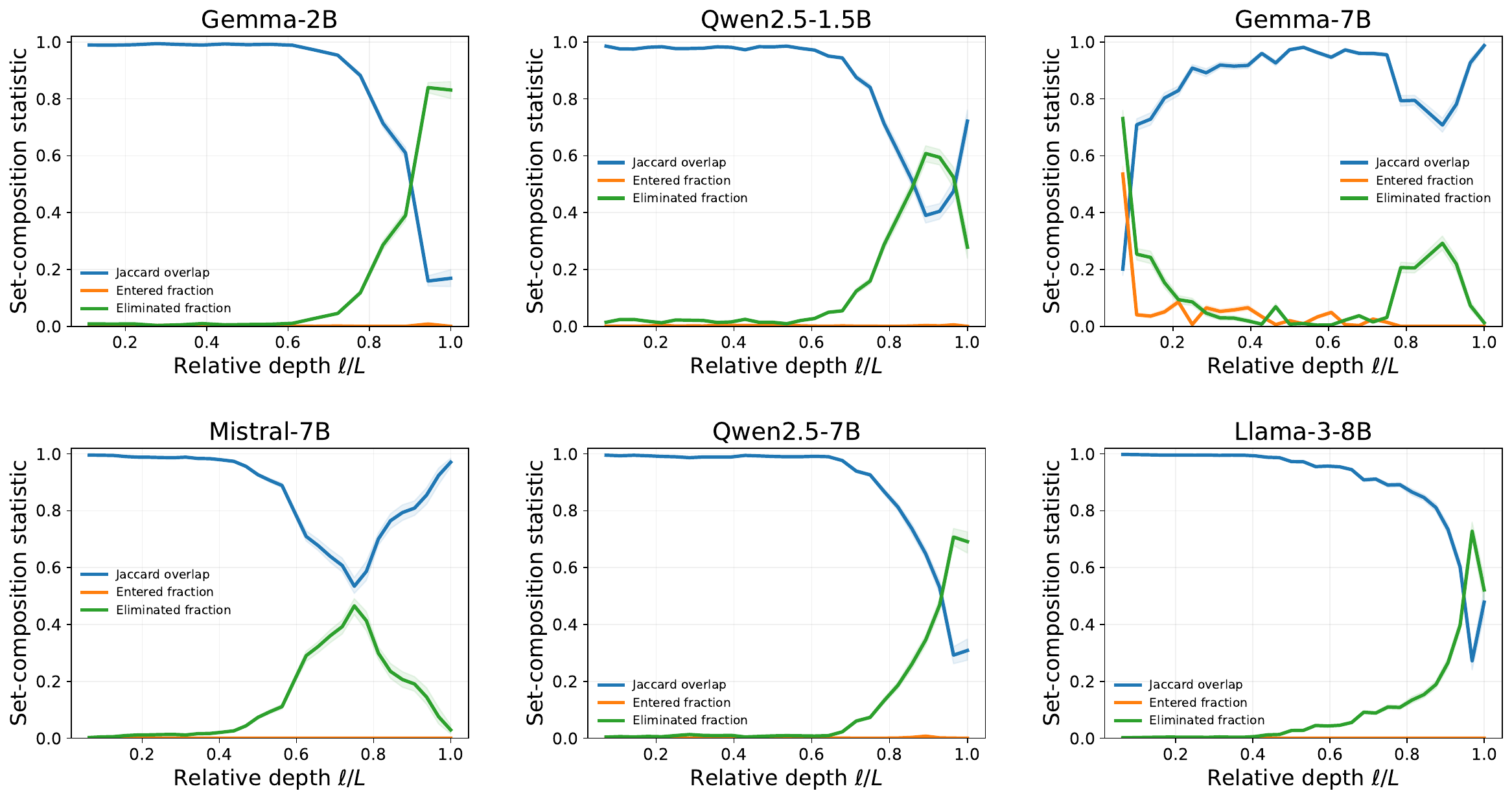}
\caption{\textbf{Euclidean competitor membership across adjacent layers.}
Mean Jaccard overlap, entry fraction, and exit fraction with document
bootstrap bands conditional on the fixed endpoint bank. Empty-to-empty
transitions contribute $J=1$.}
\label{fig:composition_euc}
\end{figure}

Euclidean entry remains near zero over much of depth, with late changes
dominated by exits (Figure~\ref{fig:composition_euc}). Cosine turnover is
more sustained in several models, especially Qwen2.5-1.5B. Resolved
entries exclude the straight-path baseline proved in
Appendix~\ref{app:marginproof}, while the statistics alone do not
establish that the network explicitly represents a set of candidate endpoints.

\section{Endpoint Geometry and Output Organization}
\label{app:modelrelevance}

Section~\ref{sec:setup} motivates residual endpoints as a
prediction-related reference population. We supplement the main
rank-conditioned cosine result with output-distribution divergence,
within-token compactness, and the Euclidean rank comparison. These
analyses assess how endpoint geometry relates to outputs without
identifying geometric competition with token probability.

\subsection{Distance and output-distribution divergence}

\begin{figure}[t]
\centering
\includegraphics[width=0.94\linewidth]{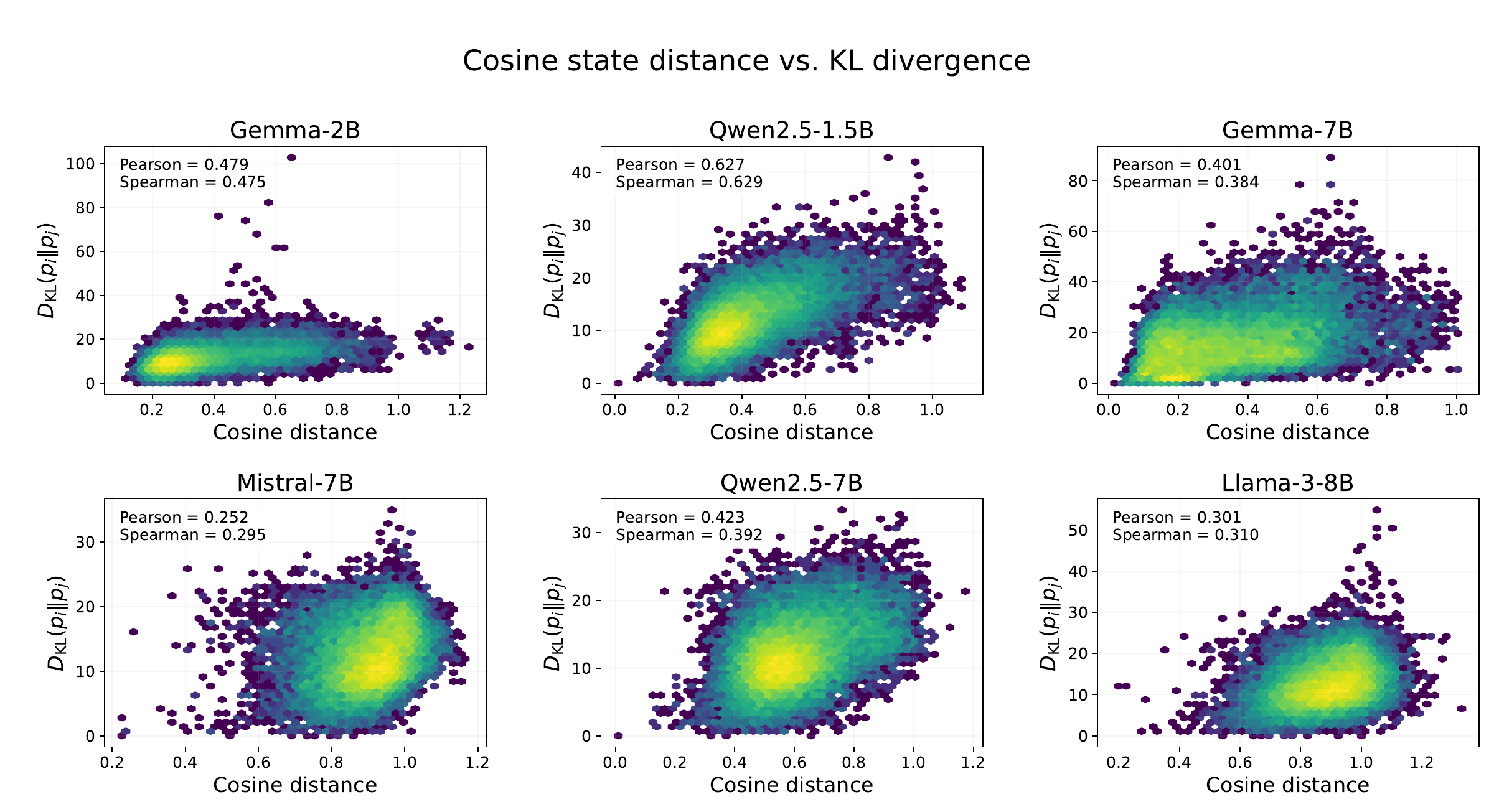}\par
{\small\textbf{(a) Cosine distance}}\par\smallskip
\includegraphics[width=0.94\linewidth]{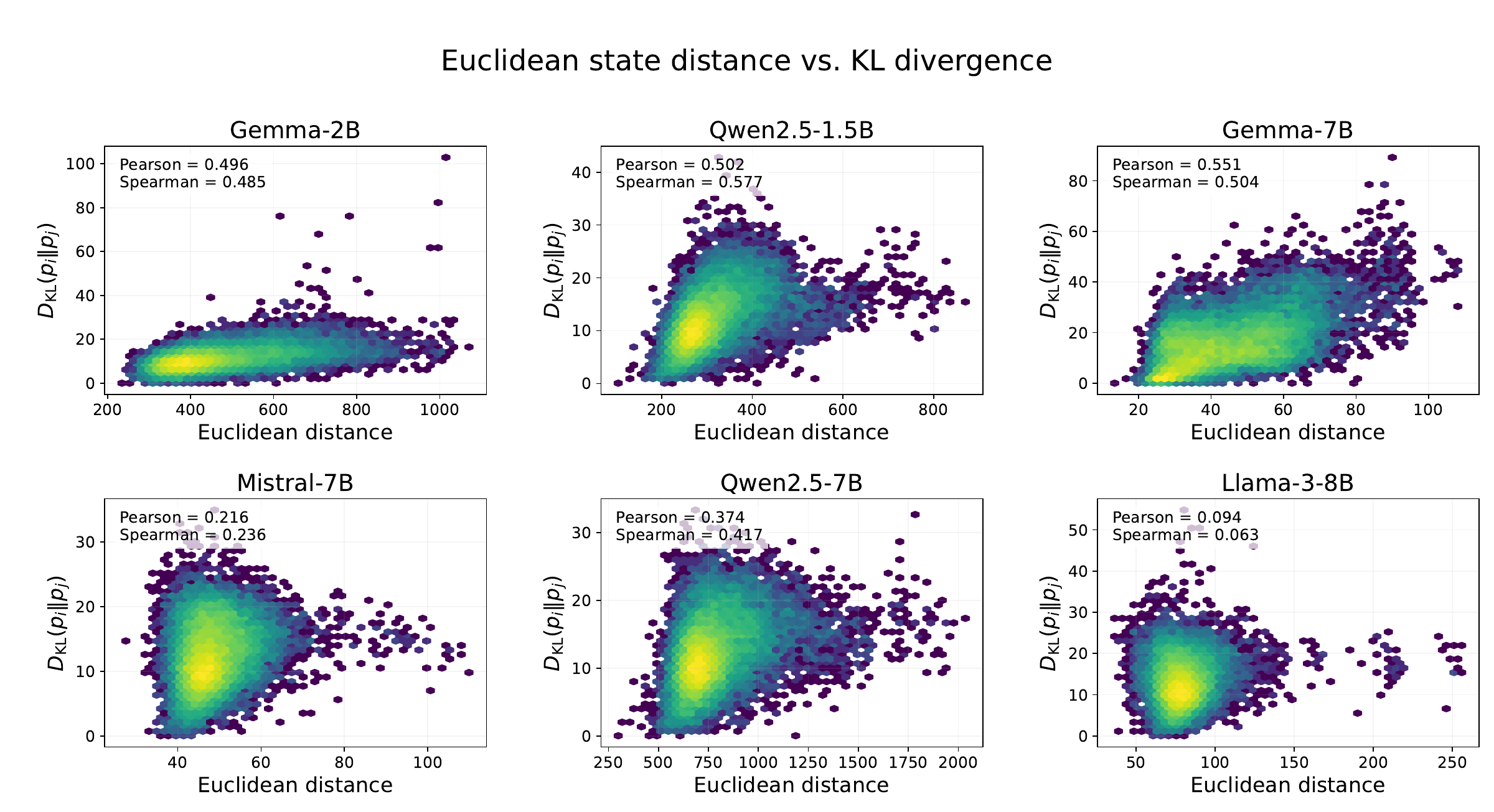}\par
{\small\textbf{(b) Euclidean distance}}
\caption{\textbf{Endpoint distance and output-distribution divergence.}
Cosine (a) and raw Euclidean (b) distance between final residual
states, compared with KL divergence between their next-token distributions.
Each panel reports Pearson and Spearman correlations.}
\label{fig:kl}
\end{figure}

Let $p_i$ be the final next-token distribution for context $i$, as defined
in Section~\ref{sec:setup}. Pairwise endpoint distance is positively
associated with the directed divergence $D_{\rm KL}(p_i\Vert p_j)$
(Figure~\ref{fig:kl}). The Pearson correlations for cosine distance range
from $0.25$ in Mistral-7B to $0.63$ in Qwen2.5-1.5B. The association
supports the use of endpoint geometry as a prediction-related probe,
although substantial variation in output divergence remains unexplained.

\subsection{Compactness of endpoints sharing a top prediction}

\begin{figure}[t]
\centering
\includegraphics[width=\linewidth]{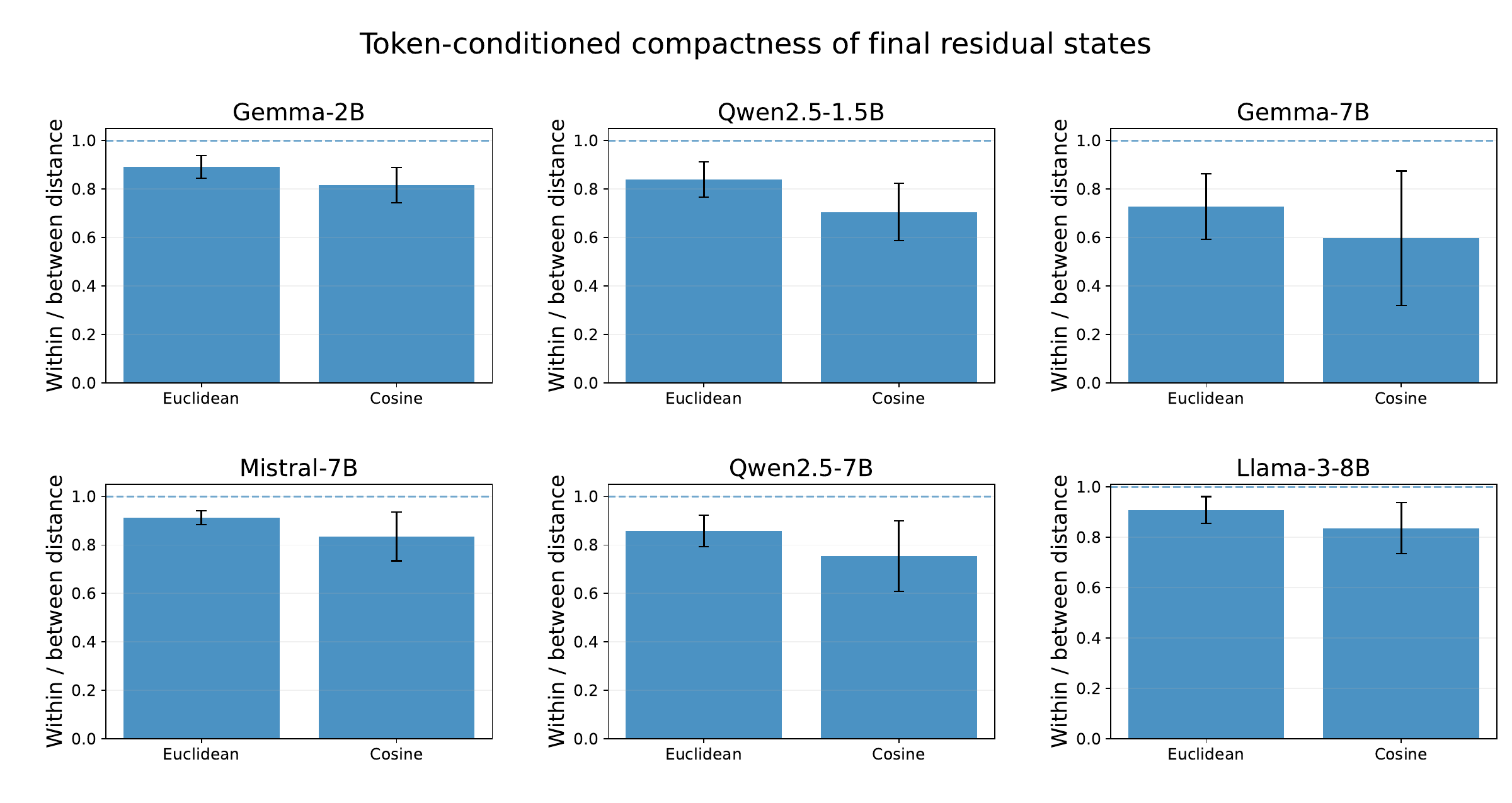}
\caption{\textbf{Compactness of endpoints sharing a top predicted token.}
Mean within-token distance divided by mean between-token distance.
Ratios below one indicate average geometric organization by the final
top prediction under both metrics.}
\label{fig:tokencompact}
\end{figure}

We group endpoints by their final most probable next token and compare
mean distances within and between token groups. The within-to-between
ratio is below one under both metrics in every model
(Figure~\ref{fig:tokencompact}). Endpoints sharing a top prediction are
therefore more compact on average. This aggregate difference does not
establish disjoint classes or a one-to-one correspondence between tokens
and endpoints. The bank retains variation among contexts with the same
top prediction, as required by the endpoint-based probe.

\subsection{Output-token rank and Euclidean endpoint distance}

For the rank-conditioned analysis in Eq.~\ref{eq:rankdist}, a query
contributes at rank $k$ only when the bank contains an alternative
endpoint whose top prediction is the query's rank-$k$ token.
Rank-specific averages can therefore involve different subsets of
queries. The curves describe represented token groups and do not assign
an endpoint distribution to every vocabulary token.

\begin{figure}[t]
\centering
\includegraphics[width=\linewidth]{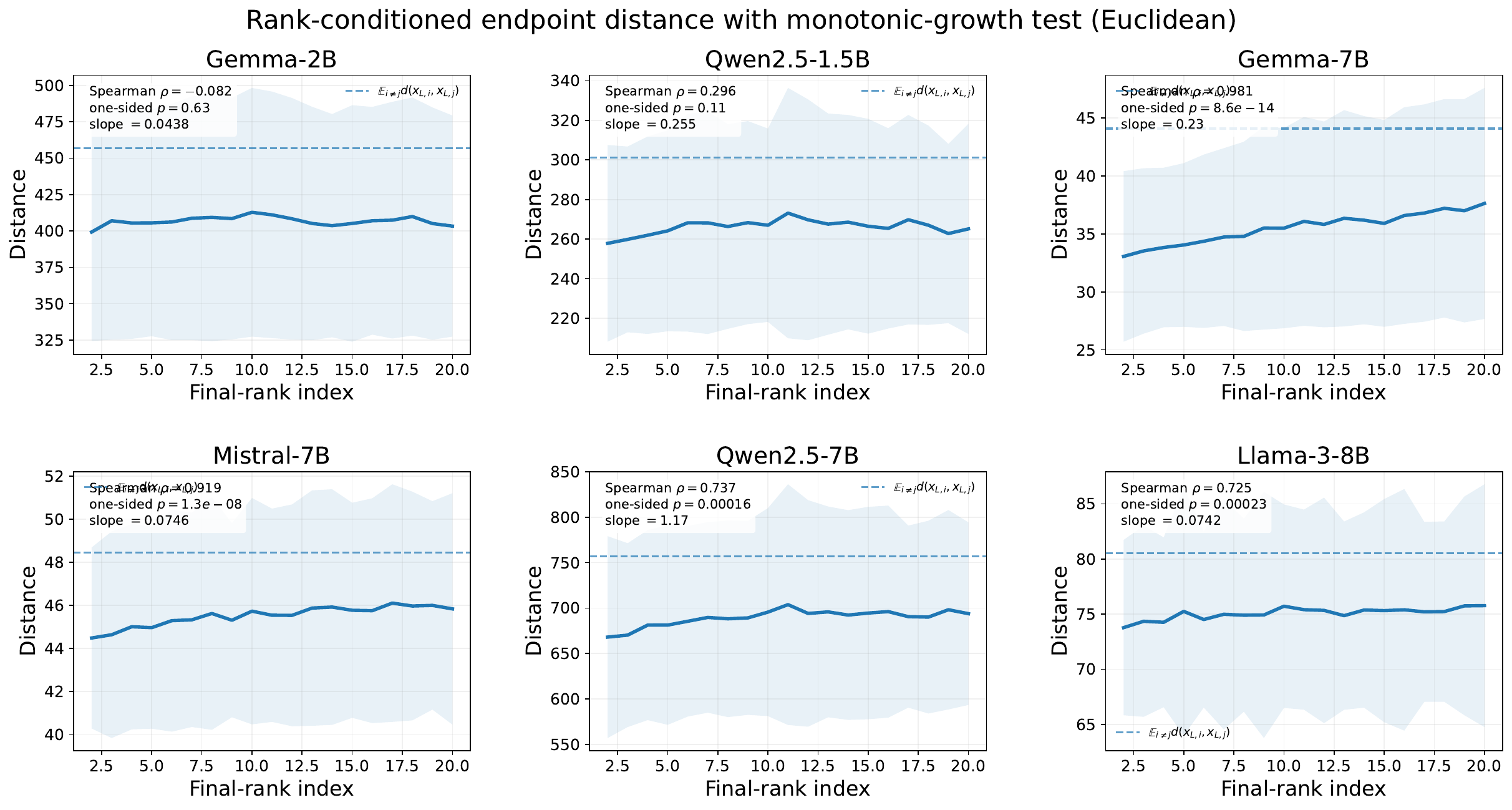}
\caption{\textbf{Euclidean endpoint distance by output-token rank.}
Mean raw Euclidean distance to endpoints whose top prediction is the
query's rank-$k$ token. Empty token groups are excluded separately at each
rank. Horizontal lines show mean distance between distinct endpoints.}
\label{fig:rank_euc}
\end{figure}

The Euclidean companion in Figure~\ref{fig:rank_euc} shows a less
consistent trend than the cosine result in Figure~\ref{fig:rank}. The
reported one-sided positive-trend tests are significant for four models,
but not for Gemma-2B ($p=0.63$) or Qwen2.5-1.5B ($p=0.11$).
The main claim about lower-probability tokens corresponding to more
distant endpoints is consequently most consistent under cosine distance.

\section{Population Geometry and Endpoint-Model Evaluation}
\label{app:models}

The endpoint-distribution experiment in Section~\ref{sec:theory} asks how
population geometry determines competition along a fixed trajectory.
We first describe the observed population spread, then report model
selection and held-out endpoint fit. The cosine prediction table connects
these distributional comparisons to the competitor curves in the main text.

\subsection{Layerwise population geometry}
\label{sec:exp_ambient}\label{app:ambient}

\begin{figure}[t]
\centering
\begin{minipage}[t]{0.49\linewidth}\centering\includegraphics[width=\linewidth]{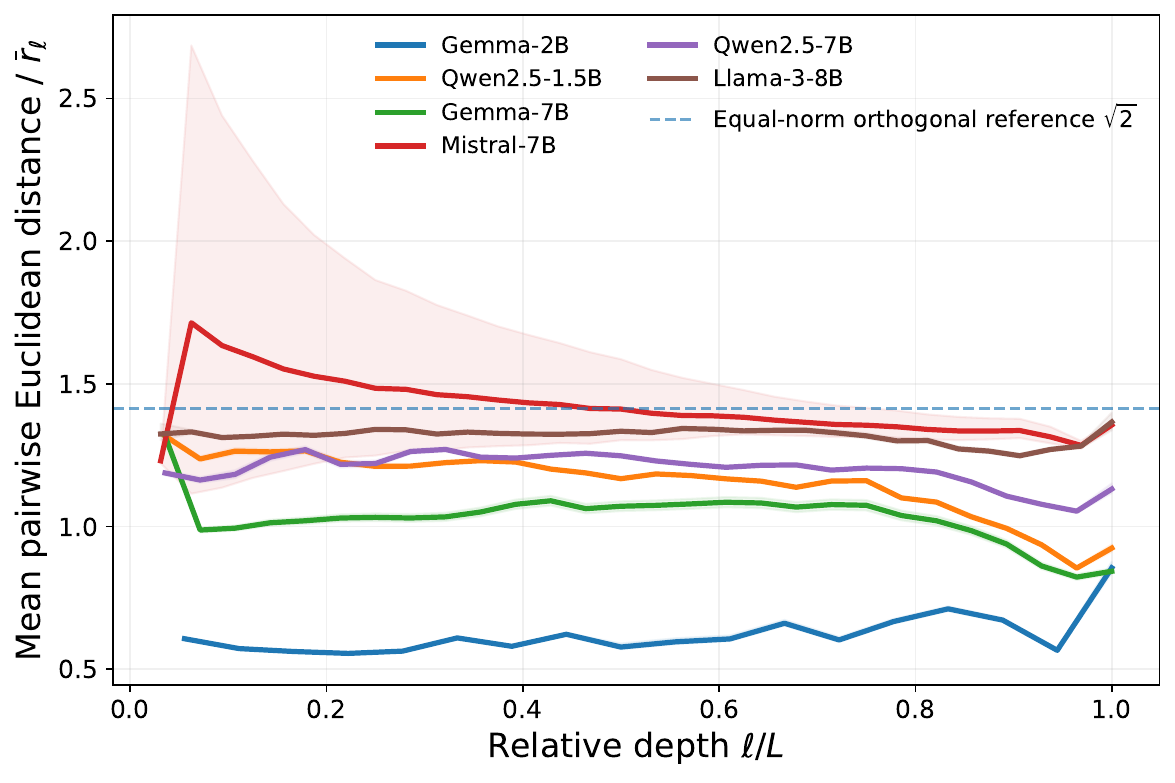}\end{minipage}\hfill
\begin{minipage}[t]{0.49\linewidth}\centering\includegraphics[width=\linewidth]{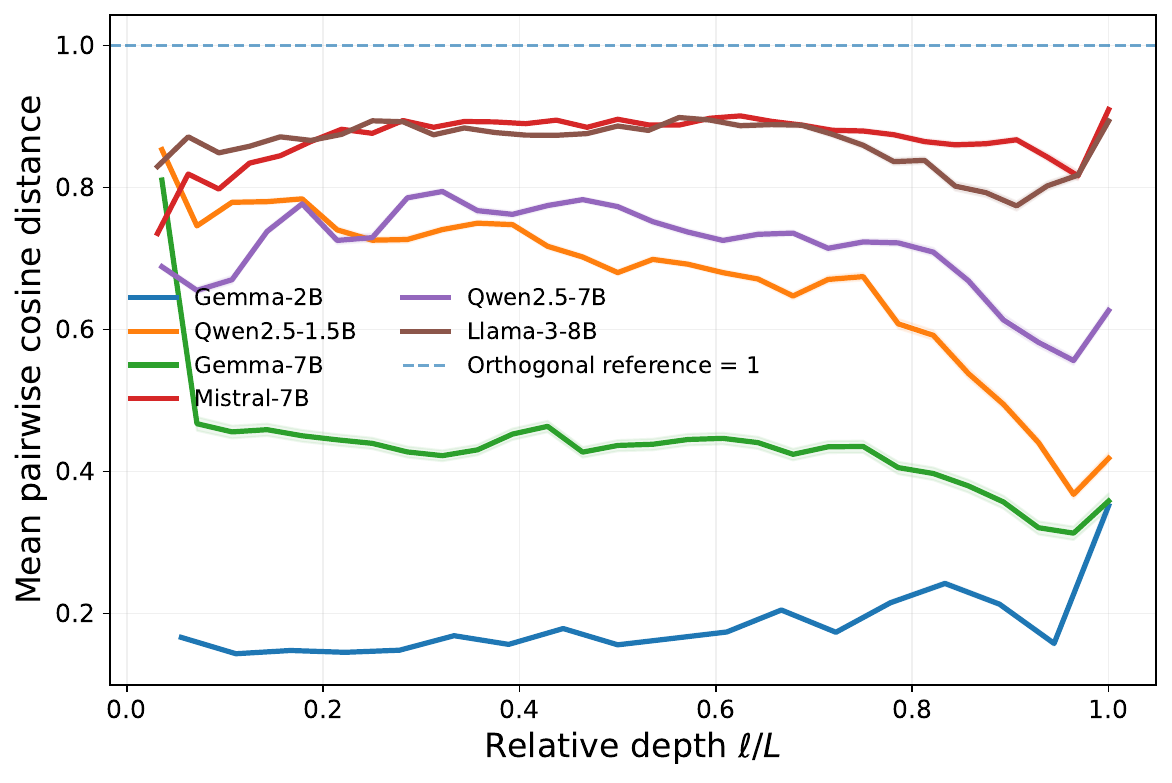}\end{minipage}
\caption{\textbf{Layerwise residual population geometry.}
Normalized mean pairwise Euclidean distance (left) and mean pairwise
cosine distance (right). Dashed references indicate equal-norm
orthogonality and zero mean directional alignment, respectively.}
\label{fig:ambient}
\end{figure}

Whole-layer geometry provides context for the directional endpoint
models. Let $\bar r_\ell=\EE_i\norm{x_{\ell,i}}$, with empirical averages
over the measured contexts. Define
\begin{equation}
 G_E(\ell)=\frac{\EE_{i\ne j}\norm{x_{\ell,i}-x_{\ell,j}}}{\bar r_\ell},
 \qquad
 G_C(\ell)=\EE_{i\ne j}[1-\cos(x_{\ell,i},x_{\ell,j})].
 \label{eq:cloud}
\end{equation}
Figure~\ref{fig:ambient} shows positive mean directional alignment and
architecture-dependent normalized spread. The reference $G_C=1$
corresponds to zero mean cosine similarity. The reference $G_E=\sqrt2$
corresponds to equal-norm orthogonal states, and requires that norm
assumption for its interpretation.

The triangle inequality gives $0\leq G_E\leq2$. Its bootstrap bands
resample precomputed context summaries with the original normalizer
$\bar r_\ell$ held fixed, so they do not estimate the full uncertainty
of the ratio. These population means motivate structured directional
models but do not determine the number of modes, concentration of
pairwise distances, or cohesion of a query's competitor set. The
construction in Appendix~\ref{app:shared_component} shows explicitly
how population spread can remain fixed while competition decreases.

\subsection{Complexity selection and held-out endpoint fit}
\label{sec:models_full}

\begin{figure}[t]
\centering
\includegraphics[width=\linewidth]{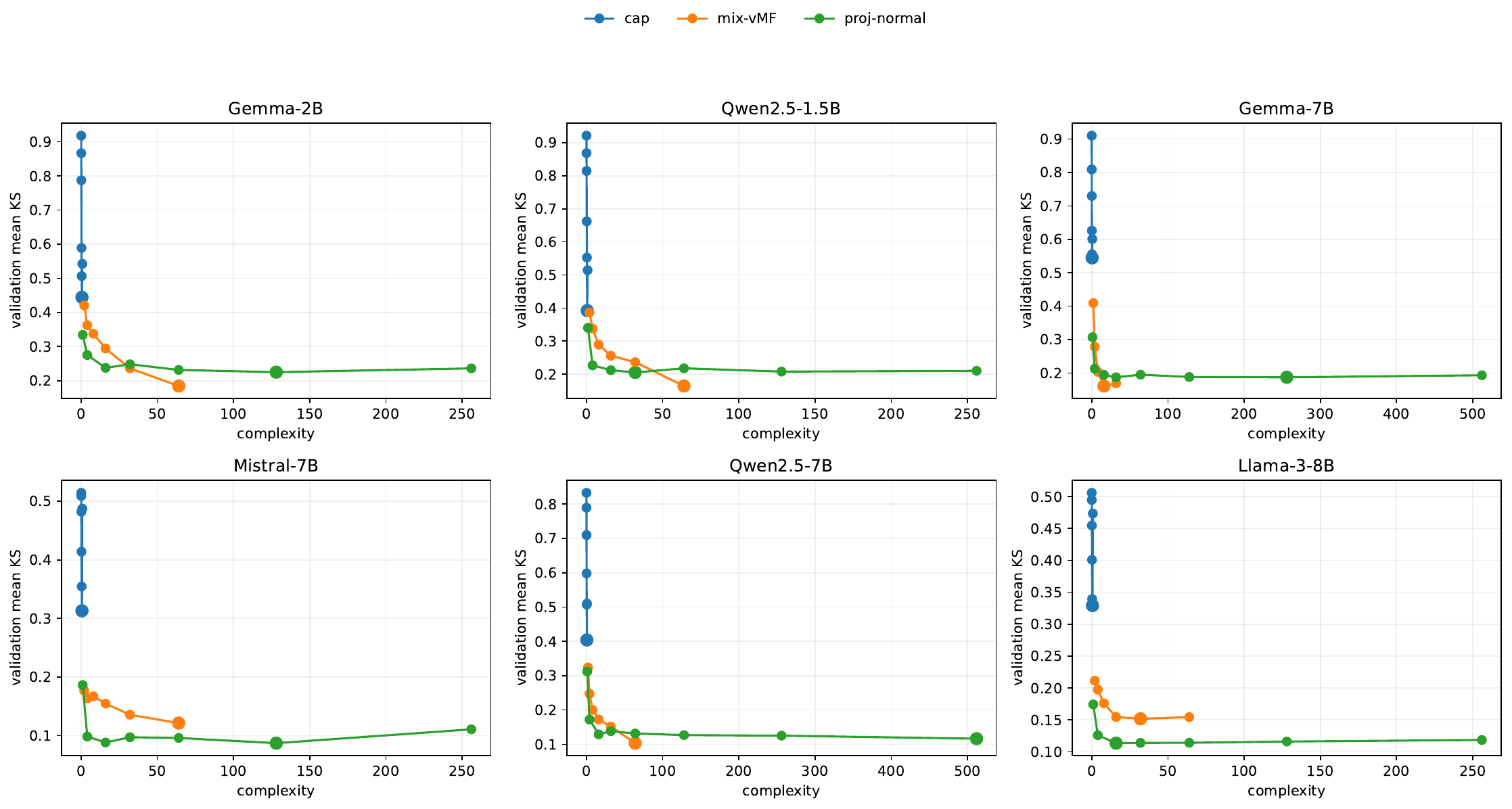}
\caption{\textbf{Validation search over directional model complexity.}
The score averages four KS distances, with smaller values indicating
closer agreement. The horizontal coordinate is the boundary quantile
$q$ for the cap, component count $C$ for the mixture, and rank $k$ for
projected normal. The small cap quantiles lie near the origin on this
shared axis.}
\label{fig:complexity_app}
\end{figure}

The validation search in Figure~\ref{fig:complexity_app} applies the
matched diagnostic score from Appendix~\ref{sec:methods_validation}.
Several mixture curves continue to improve through $C=64$, and
Qwen2.5-7B selects the largest tested projected-normal rank, $k=512$.
These boundary optima leave open whether higher complexity would improve
the corresponding fits. All reported configurations are fixed before
test evaluation.

Table~\ref{tab:globalfit} identifies the lowest test diagnostic score
among the validation-selected families for each model. For
Qwen2.5-1.5B, projected normal and the mixture are nearly tied at
$0.151$ and $0.153$. The best reported scores improve on the sphere by
factors of $4.6$--$7.5$. These diagnostics assess endpoint samples, while
trajectory prediction separately assesses mass in query-dependent
comparison regions. The family with the best endpoint diagnostic score
need not have the smallest trajectory-prediction error.

\begin{table}[t]
\centering
\small
\begin{tabular}{llrr}
\toprule
Model & Lowest-score family & Complexity & Test score \\
\midrule
Gemma-2B & vMF mixture & $C=64$ & 0.206 \\
Qwen2.5-1.5B & Projected normal & $k=32$ & 0.151 \\
Gemma-7B & Projected normal & $k=256$ & 0.192 \\
Mistral-7B & Projected normal & $k=128$ & 0.079 \\
Qwen2.5-7B & vMF mixture & $C=64$ & 0.116 \\
Llama-3-8B & Projected normal & $k=16$ & 0.094 \\
\bottomrule
\end{tabular}
\caption{\textbf{Lowest held-out endpoint diagnostic scores.}
Complexity is selected within each family on validation documents.
The table reports the smallest resulting test score for each language
model as a descriptive comparison among families.}
\label{tab:globalfit}
\end{table}

\subsection{Cosine competitor-fraction prediction}
\label{app:cosine_prediction}

Table~\ref{tab:traj_log} quantifies the curves in Figure~\ref{fig:predicted}
using the preterminal log-scale error. Projected normal has the lowest
synthetic point estimate in five models, and the vMF mixture is slightly
better in Gemma-7B. Both families outperform the cap, single vMF, and
sphere in all six models. The paired comparisons in
Appendix~\ref{app:prediction} distinguish projected normal from the
mixture in three models. The results support the contribution of
directional structure to competitive mass, without establishing one
family as uniformly superior across all comparisons.

\begin{table}[t]
\centering
\small
\setlength{\tabcolsep}{3.2pt}
\begin{tabular}{lrrrrrr}
\toprule
Model & Real & Sphere & Cap & vMF & Mix-vMF & Proj.-normal \\
\midrule
Gemma-2B      & 0.040 & 0.537 & 0.379 & 0.329 & 0.086 & \textbf{0.033} \\
Qwen2.5-1.5B  & 0.101 & 1.278 & 0.365 & 0.429 & 0.195 & \textbf{0.117} \\
Gemma-7B      & 0.020 & 1.910 & 0.266 & 0.348 & \textbf{0.102} & 0.103 \\
Mistral-7B    & 0.019 & 0.160 & 0.080 & 0.077 & 0.023 & \textbf{0.022} \\
Qwen2.5-7B    & 0.084 & 1.657 & 0.573 & 0.642 & 0.186 & \textbf{0.139} \\
Llama-3-8B    & 0.035 & 0.138 & 0.150 & 0.156 & 0.108 & \textbf{0.107} \\
\bottomrule
\end{tabular}
\caption{\textbf{Cosine competitor-fraction prediction error.}
Mean absolute log-scale error in dex over layers $1,\ldots,L-1$.
Bold marks the smallest synthetic point estimate. ``Real'' is the
reference sampled from training endpoints. These point estimates are
complemented by the uncertainty analysis in Appendix~\ref{app:prediction}.}
\label{tab:traj_log}
\end{table}

\section{Sensitivity to Endpoint-Bank Size}
\label{app:banksize}

The competitor fraction in Section~\ref{sec:turnover} is normalized to
support comparison across eligible bank sizes. For a fixed query and a
bank of $M_i>1$ alternatives, let $K_i=|\CC_{\ell,i}|$. If $1\leq s\leq M_i$ alternatives
are sampled uniformly without replacement, their competitor count
$K_{i,s}$ has a hypergeometric distribution. Consequently,
\begin{equation}
 \EE\!\left[\frac{K_{i,s}}{s}\right]=\frac{K_i}{M_i},
 \qquad
 \operatorname{Var}\!\left(\frac{K_{i,s}}{s}\right)
 =\frac{A_{\ell,i}(1-A_{\ell,i})}{s}
 \frac{M_i-s}{M_i-1}.
 \label{eq:app_subsample}
\end{equation}
Thus uniform subsampling preserves the expected fraction for the fixed
query while increasing sampling variation and coarsening the fraction's
resolution to $1/s$.

\begin{figure}[t]
\centering
\includegraphics[width=\linewidth,trim=0 0 0 88pt,clip]{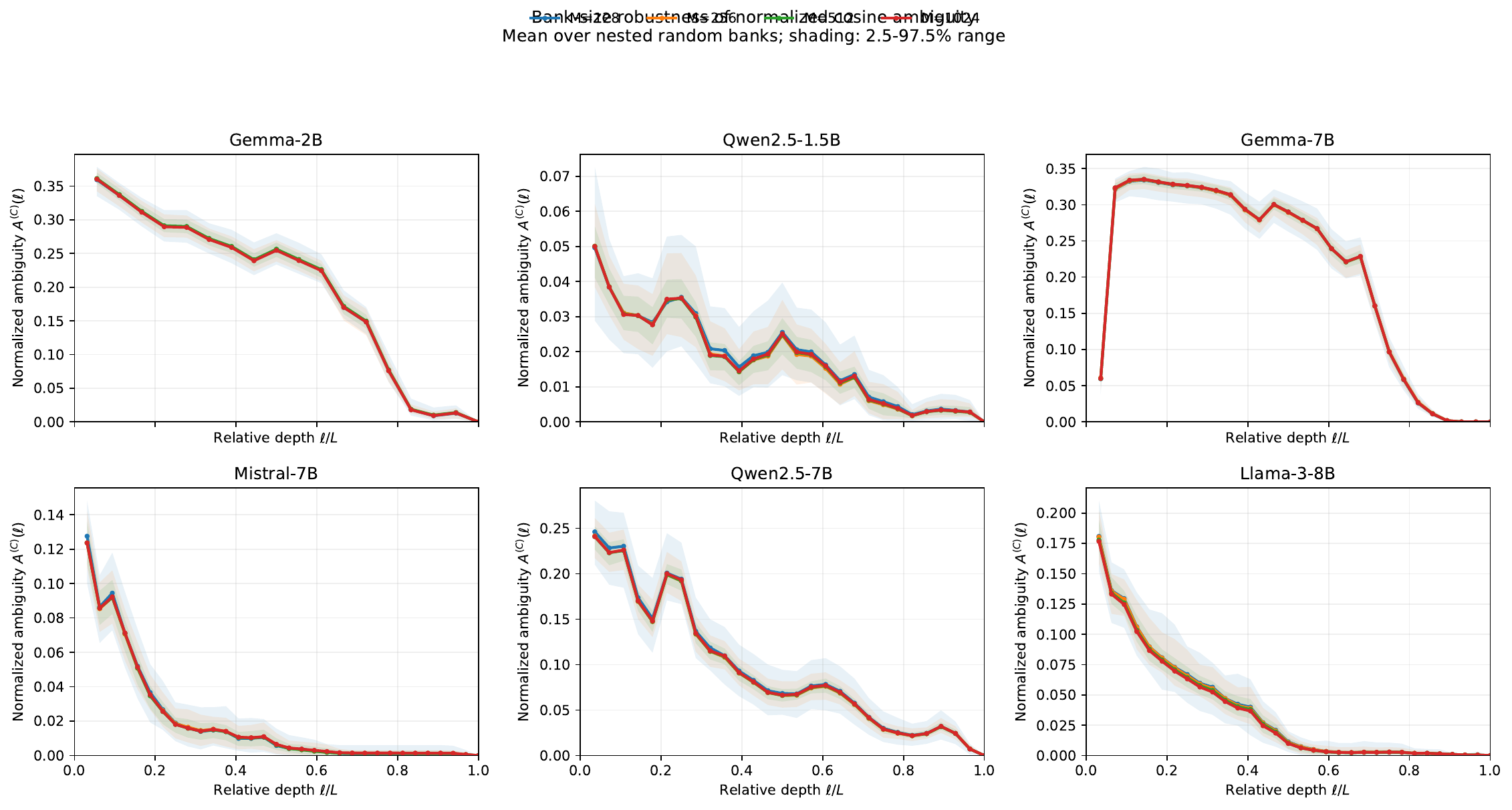}\par
\smallskip
{\small\textcolor[rgb]{0.122,0.467,0.706}{\rule{1em}{1.2pt}} $n_{\rm bank}=128$
\quad\textcolor[rgb]{1.000,0.498,0.055}{\rule{1em}{1.2pt}} $256$
\quad\textcolor[rgb]{0.173,0.627,0.173}{\rule{1em}{1.2pt}} $512$
\quad\textcolor[rgb]{0.839,0.153,0.157}{\rule{1em}{1.2pt}} $1{,}024$}
\caption{\textbf{Cosine competitor fractions across bank sizes.}
Mean curves for nested samples of 128, 256, 512, and 1,024 contexts.
Shading gives the 2.5--97.5\% range across bank draws. The vertical-axis
label ``normalized ambiguity'' denotes the competitor fraction
$A^{(C)}(\ell)$ used throughout the paper.}
\label{fig:banksize}
\end{figure}

Figure~\ref{fig:banksize} evaluates nested random samples containing
$n_{\rm bank}\in\{128,256,512,1{,}024\}$ contexts. These values count
sampled contexts. Under Eq.~\ref{eq:cset}, the eligible alternative count excludes the query's
own endpoint whenever it is present, and is therefore
$M_i=n_{\rm bank}-\ind\{i\text{ is in the sampled bank}\}$.
In particular, the full 1,024-context bank provides 1,023 alternatives.

Mean cosine curves overlap closely. The largest visible differences
occur at early depth in Mistral-7B and Qwen2.5-7B and at intermediate
depth in Qwen2.5-1.5B. Their absolute size is below $0.01$ and within
the sampling range of the smallest bank. The main depth patterns,
including the late increase in Qwen2.5-7B, persist across samples.
This supports stability within the available endpoint population.
Transfer across documents is assessed separately by the endpoint-model
experiment. The bank-size check was performed only for cosine distance.

\section{Additional Trajectory-Prediction Results}
\label{app:prediction}

Section~\ref{sec:theory} emphasizes cosine prediction because it directly
tests the fitted directional structure. We report the Euclidean
comparison to assess the role of endpoint norms and give the available
uncertainty summaries for the cosine errors.

\subsection{Euclidean competitor-fraction prediction}

\begin{figure}[t]
\centering
\includegraphics[width=\linewidth]{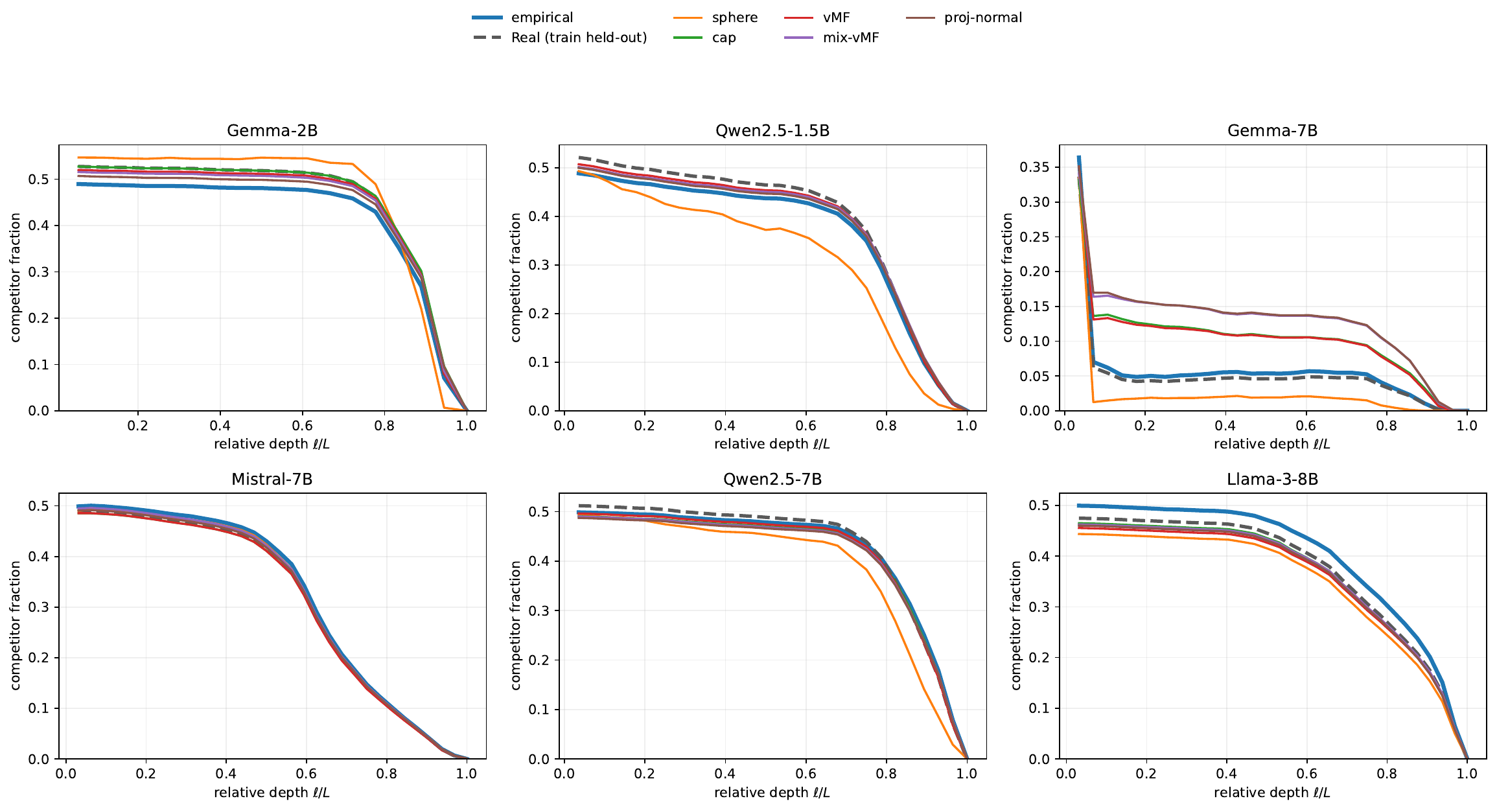}
\caption{\textbf{Predicting Euclidean competitor fractions.}
Empirical test fractions, predictions from fitted endpoint distributions,
and the training-endpoint reference. Sampled banks remain fixed across
layers, and predictions are averaged over eight banks.}
\label{fig:predicted_euc}
\end{figure}

Figure~\ref{fig:predicted_euc} and Table~\ref{tab:traj_euc} use the same
held-out trajectories, sampled banks, and preterminal error as the
cosine analysis. Family rankings vary across architectures. Projected
normal has the smallest error for Gemma-2B, single vMF for Gemma-7B,
and the mixture for Mistral-7B. Other comparisons are tied at the
reported precision, including cap and projected normal for
Qwen2.5-1.5B, cap and vMF for Qwen2.5-7B, and cap and mixture for
Llama-3-8B.

The Euclidean results do not identify a consistently preferred family.
Euclidean membership depends jointly on endpoint norms and directions,
as shown in Appendix~\ref{app:shared_component}, and all fitted families
assume independence between these quantities. Several synthetic errors
are below the finite real-bank reference, consistent with that
reference's role as a sampling comparison.

\begin{table}[t]
\centering
\small
\setlength{\tabcolsep}{3.2pt}
\begin{tabular}{lrrrrrr}
\toprule
Model & Real & Sphere & Cap & vMF & Mix-vMF & Proj.-normal \\
\midrule
Gemma-2B      & 0.034 & 0.109 & 0.040 & 0.029 & 0.027 & \textbf{0.023} \\
Qwen2.5-1.5B  & 0.026 & 0.128 & \textbf{0.015} & 0.018 & 0.018 & \textbf{0.015} \\
Gemma-7B      & 0.053 & 0.490 & 0.314 & \textbf{0.299} & 0.414 & 0.420 \\
Mistral-7B    & 0.013 & 0.012 & 0.018 & 0.021 & \textbf{0.006} & 0.010 \\
Qwen2.5-7B    & 0.012 & 0.067 & \textbf{0.009} & \textbf{0.009} & 0.010 & 0.013 \\
Llama-3-8B    & 0.033 & 0.069 & \textbf{0.043} & 0.050 & \textbf{0.043} & 0.047 \\
\bottomrule
\end{tabular}
\caption{\textbf{Euclidean competitor-fraction prediction error in dex.}
The error follows Eq.~\ref{eq:methods_logerr}. Bold marks the smallest
synthetic value at the reported precision, including ties after rounding.}
\label{tab:traj_euc}
\end{table}

\subsection{Cosine uncertainty and paired comparisons}

Table~\ref{tab:prediction_uncertainty} gives the 95\% document-bootstrap
intervals for the real-reference errors and summarizes the paired
comparison between projected normal and the vMF mixture. The paired
analysis distinguishes the two fitted families for Gemma-2B and both
Qwen models, favoring projected normal. It does not distinguish them
for Gemma-7B, Mistral-7B, or Llama-3-8B. Small differences between the
point estimates in Table~\ref{tab:traj_log} should be interpreted in
light of this uncertainty.

\begin{table}[t]
\centering
\small
\begin{tabular}{lrl}
\toprule
Model & Real-reference error: 95\% interval & Paired PN--mixture result \\
\midrule
Gemma-2B & $[0.034,0.049]$ & favors projected normal \\
Qwen2.5-1.5B & $[0.083,0.127]$ & favors projected normal \\
Gemma-7B & $[0.018,0.024]$ & not distinguished \\
Mistral-7B & $[0.014,0.030]$ & not distinguished \\
Qwen2.5-7B & $[0.076,0.091]$ & favors projected normal \\
Llama-3-8B & $[0.028,0.050]$ & not distinguished \\
\bottomrule
\end{tabular}
\caption{\textbf{Uncertainty in cosine trajectory-prediction error.}
Intervals and paired comparisons use 1,000 resamples of test documents,
conditional on fitted distributions and sampled banks. PN denotes
projected normal. The reported intervals concern the real-bank reference,
not the difference between the fitted families.}
\label{tab:prediction_uncertainty}
\end{table}

The paired comparison uses the same query-document weights for both
families, as specified in Appendix~\ref{sec:methods_prediction}. It
therefore controls for differences in the resampled query population,
while leaving uncertainty from endpoint-model fitting and new bank draws
outside the comparison. Numerical intervals for the paired error
differences are not available in the accompanying source materials.

\FloatBarrier
\section{Reproducibility and Source Materials}
\label{app:repro}

Reproducing the paper requires preserving the reference population for
each experiment, because competitor fractions depend on both the query
trajectory and its eligible endpoint bank. The primary geometry analysis
uses 1,024 contexts per model, the matched input-token control uses a
separate 2,000-context sample, and endpoint fitting uses disjoint document
partitions of the primary sample. These protocols are specified in
Appendices~\ref{app:protocol} and~\ref{app:tokencontrols}.

The accompanying project contains manuscript sources and figure PDFs.
Extraction and analysis code, per-context residual arrays, fitted
endpoint models, and random seeds are not included. These materials are
needed to independently reproduce the reported numerical results and
resolve the implementation details identified in source comments,
including the alternative-endpoint projection convention, the
projected-normal sampling implementation, the bank-size sampling
convention, and the reported rank-trend test.

\subsection*{AI use statement}
Language-model tools were used for language editing, restructuring,
mathematical derivation, and source debugging. The authors are responsible
for the mathematical statements, experimental design, data analysis,
citations, and final text.

\end{document}